\documentclass[APA,Times1COL]{WileyNJDv5} %STIX1COL,STIX2COL,STIXSMALL

\articletype{Journal of Field Robotics}%

\received{28 June 2023}
\revised{19 March 2024}
\accepted{21 March 2024}
\journal{Journal of Field Robotics}
\volume{00}
\copyyear{2024}
\startpage{1}

\usepackage{color}
\usepackage{colortbl}
\begin{document}

\title{Motion planning for off-road autonomous driving based on human-like cognition and weight adaptation}

\author[1]{Yuchun Wang}

\author[1]{Cheng Gong}

\author[1]{Jianwei Gong}

\author[1]{Peng Jia}

\authormark{Yuchun Wang\textsc{et al.}}
\titlemark{Motion planning for off-road autonomous driving based on human-like cognition and weight adaptation}

\address[1]{\orgdiv{The authors are with the School of Mechanical Engineering}, \orgname{Beijing Institute of Technology}, \orgaddress{\state{Beijing 100081}, \country{China}}}

\corres{Corresponding author Jianwei Gong, \email{gongjianwei@bit.edu.cn} }

%\fundingInfo{Text}
%\JELinfo{ejlje}

\abstract[Abstract]{
Driving in an off-road environment is challenging for autonomous vehicles due to the complex and varied terrain. To ensure stable and efficient travel, the vehicle requires consideration and balancing of environmental factors, such as undulations, roughness, and obstacles, to generate optimal trajectories that can adapt to changing scenarios. However, traditional motion planners often utilize a fixed cost function for trajectory optimization, making it difficult to adapt to different driving strategies in challenging irregular terrains and uncommon scenarios. To address these issues, we propose an adaptive motion planner based on human-like cognition and cost evaluation for off-road driving. First, we construct a multi-layer map describing different features of off-road terrains, including terrain elevation, roughness, obstacle, and artificial potential field map. Subsequently, we employ a CNN-LSTM network to learn the trajectories planned by human drivers in various off-road scenarios. Then, based on human-like generated trajectories in different environments, we design a primitive-based trajectory planner that aims to mimic human trajectories and cost weight selection, generating trajectories that are consistent with the dynamics of off-road vehicles. Finally, we compute optimal cost weights and select and extend behavioral primitives to generate highly adaptive, stable, and efficient trajectories. 

We validate the effectiveness of the proposed method through experiments in a desert off-road environment with complex terrain and varying road conditions. The experimental results show that the proposed human-like motion planner has excellent adaptability to different off-road conditions. It shows real-time operation, greater stability, and more human-like planning ability in diverse and challenging scenarios.
	}


\jnlcitation{\cname{%
\author{Taylor M.},
\author{Lauritzen P},
\author{Erath C}, and
\author{Mittal R}}.
\ctitle{On simplifying ‘incremental remap’-based transport schemes.} \cjournal{\it J Comput Phys.} \cvol{2021;00(00):1--18}.}

\maketitle

%\renewcommand\thefootnote{}
%\footnotetext{\textbf{Abbreviations:} ANA, anti-nuclear antibodies; APC, antigen-presenting cells; IRF, interferon regulatory factor.}

%\renewcommand\thefootnote{\fnsymbol{footnote}}
%\setcounter{footnote}{1}

\section{Introduction}

Off-road driving is a significant challenge for autonomous vehicles as it requires navigating through various complex scenarios such as ramps, rough terrain, undulations, unstructured intersections. Due to its unstructured  nature, off-road environment can be difficult to model and represent, where differences in scenarios can vary in many different factors. Dense representation may be flooded with redundant information while sparse representation may missing key features. Additionally, different driving strategies should be adopted considering different metrics like safety, stability, and efficiency in various scenarios. For instance, in flat and traversable terrain, driving efficiency should be prioritized. But in rough terrain, a less rough trajectory should be chosen in order to secure safety and driving performance, where driving stability should weight more than travel efficiency. Such adaption can be difficult for traditional off-road motion planners that use predefined combinations of metrics, where the optimal combinations can hardly be estimation for all scenarios.

\begin{figure}[!t]
	\centering
	\includegraphics[width=6.8in]{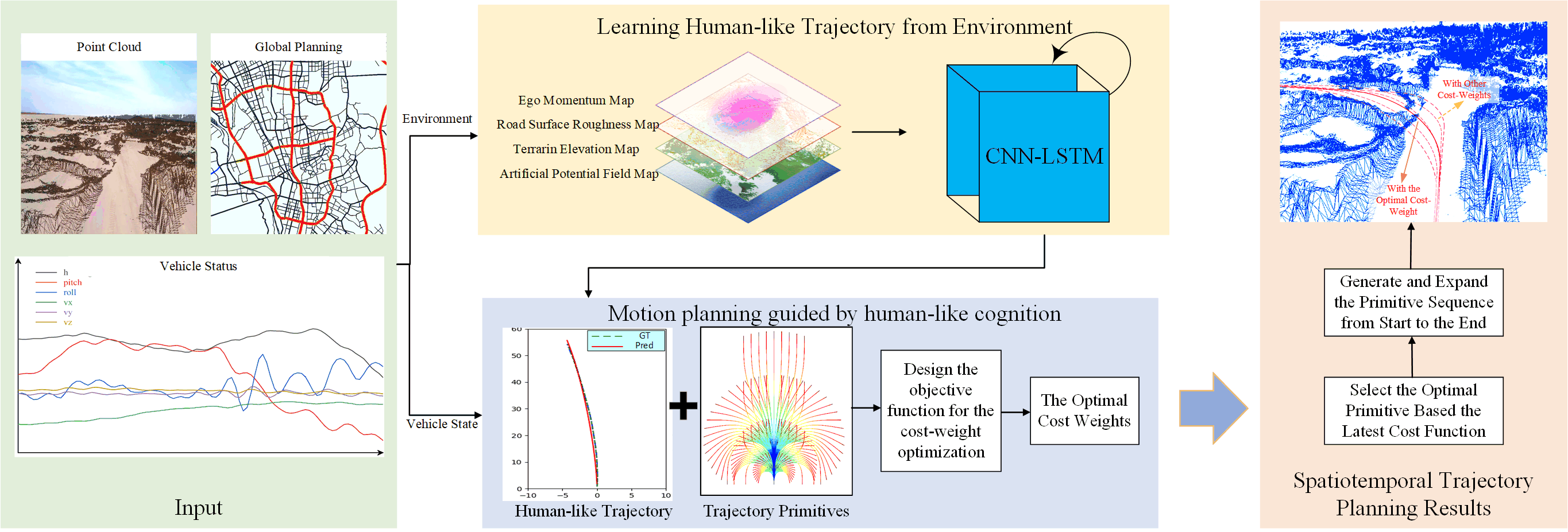}
	\caption{Flowchat. The left green box contains environmental point cloud data, global planning, and vehicle state. These inputs are fed into the human-like environment perception and learning framework located in the upper middle yellow box. This framework generates a traversal cost map and learns human-like trajectories using a CNN-LSTM network. The human-like results are then fed into the motion planning framework based on human-like cognition represented by the lower middle purple box, where the optimal cost weights are determined. Finally, these inputs are processed by the spatio-temporal coupling planning framework located in the right red box, resulting in the trajectory planning results with the optimal traversal cost weights.}
	\label{flowchat}
\end{figure}
In order to properly model complex off-road terrain, various methods have been proposed to describe off-road scenarios in a more accurate and comprehensive manner. Early off-road autonomous vehicles typically process the environment as a binary grid map \cite{43_lalonde2006natural,44_stentz2003real}, which considers traversability but fails to provide more in formation for efficiency, stability, and vehicle dynamics, for off-road motion planning. Some approaches directly used 3D point clouds for planning \cite{02_krusi2017driving}, but the sparse and unordered map led to the slow computation and poor interpretability. To improve the efficiency, some methods use 2.5D maps that complement elevation information in 2D maps \cite{06_liu2022efficient}, and some use roughness information as additional data \cite{04_waibel2022rough}. However, these single feature maps were not able to fully express various terrains such as undulations, roughness, ramps, and others in off-road scenarios. \cite{29_overbye2020fast} proposed to construct four types of feature maps for the environment and combine them with specific weights into one cost map. However, the combined map usually couples different features with an equal weights and fails to adapt to different types of terrain with different driving preference.

To enable planners to better adapt to different environment with different strategies, methods that considers weighting parameters of metrics are also studied. Early off-road motion planners focused primarily on trajectory traversability \cite{42_team2005stanford,45_pivtoraiko2011kinodynamic}, with less consideration given to factors such as stability, efficiency, or human-like driving style. Some recent approaches have defined their own costs \cite{10_hedrick2020terrain} or based on user requirements \cite{16_ji2018adaptive} for motion planners. While these works consider multiple costs, the weights are optimal only in specific environments. It is difficult to adapt to the new scenarios when road conditions change. The authors in \cite{04_waibel2022rough} find it necessary to adjust the weights in environments beyond the training. However, they only provided recommended cost weights and suggest that people adjust the cost weights to be optimal in new road conditions.

In comparison, human drivers have rich driving experience and flexible environmental cognition, which enables them to consider multiple factors in different off-road scenarios. Therefore, to address the problems of insufficient expression of off-road terrain and insensitivity to changes in road conditions in the above research, we propose a motion planning method that considers the adaptation of traversal cost weights by learning from human driving experience for off-road driving. The algorithm flowchart is shown in Fig.\ref{flowchat}. First, we construct hierarchical 2.5D feature maps using sensor and vehicle state information. Next, the maps are fed into a CNN-LSTM network to acquire knowledge about the cognitive perception of the costs associated with environmental features by the human driver. Then, behavioral primitives are generated based on the vehicle state, with the predicted trajectory from human cognition serving as the optimization objective. Finally, the optimal cost weights are solved, and the primitives are selected and extended to generate highly adaptive and human-like motion planning results. The contributions in our paper are as follows:

\begin{enumerate}
	\item{We construct hierarchical 2.5D feature maps as the traversal cost to express the complex terrain in the off-road environment.  The optimal weight in current learned from human environmental cognition of each cost adapts the variability of scenarios to generate a spatiotemporal motion planning trajectory.}
	\item{We select and extend human-like behavioral primitives based on the solved optimal traversal cost weights to generate the planning trajectory. It not only matches the trajectory characteristics of human drivers, but also satisfies the vehicle dynamics constraints, which can be directly output to the control layer for trajectory tracking.}
	\item{The effectiveness of the proposed method has been verified through experiments in a desert off-road environment. It shows real-time operation, greater stability, and more human-like planning ability in diverse and challenging scenarios, including ramps, intersections, long curves, undulations, and rough terrain.}
\end{enumerate}

The structure of this paper is as follows: Section 2 reviews related works, analyzes the advantages and disadvantages of existing methods, and identifies the problems that need to be addressed. Section 3 provides how to learn Human-like cognition with a CNN-LSTM network.Section 4 provides a detailed description of motion planning guided by human-like cognition. Section 5 presents the experimental results and analysis. Finally, the conclusion section summarizes the paper and suggests further research directions.

\section{Related Work}

%\subsection{Off-Road Environment Planning }

%\subsubsection{Environmental Processing}
\subsection{Off-Road Environment Processing}
Traditional two-dimensional (2D) maps are inadequate for representing off-road environments. One approach \cite{02_krusi2017driving} involves direct utilization of raw point clouds for planning. However, point cloud distributions lack order and are discrete, resulting in limited 3D information and slow computation. Another method normalizes the representation of the environment using 3D meshes \cite{08_ruetz2019ovpc}. Nevertheless, this approach leads to a significant number of empty meshes, causing computational inefficiencies. In contrast, \cite{16_ji2018adaptive} employs irregular voxel meshes with tree search to enhance computational speed. However, irregular voxels are unsuitable for motion planners. Consequently, the popularity of 2.5D maps, rather than full 3D maps, has grown. \cite{06_liu2022efficient} introduces an adaptive resolution 2.5D map to accommodate complex terrains in off-road scenarios. Furthermore, \cite{07_stolzle2022reconstructing} utilizes deep learning to reconstruct occluded regions and compensate for elevation information in blind spots. Additionally, \cite{09_buchanan2021perceptive} proposes a layered 2.5D elevation map approach to preserve more 3D information.

Several 2.5D maps include additional terrain information. \cite{04_waibel2022rough} explores terrain roughness through point clouds, while \cite{03_polevoy2022complex} predicts non-rigid terrain traversability by learning trajectory model errors. \cite{11_gasparino2022wayfast} directly learns the traversability of the environment to generate traversable paths, and \cite{20_dang2020graph} evaluates the performance of each off-road path to assess its traversability for vehicles. Additionally, \cite{10_hedrick2020terrain} and \cite{37_jiang2021r2} detect multiple factors in the Martian environment and determine speed limits using table lookups. However, these methods rely on recognizable and known environments. On the other hand, \cite{15_ganganath2015constraint} and \cite{12_wei2021predicting} take various environmental factors into account and generate energy cost maps using either rule-based or learned approaches. \cite{schmid2022self} proposed a method for predicting high-and low-risk terrains based on past vehicle experience. Although these methods provide cost estimates for specific scenarios, they struggle to adapt to off-road environments with complex road conditions and the incorporation of human driving preferences. In contrast, \cite{28_tian2023driving} constructs a potential field cost map based on obstacles and risks, while \cite{29_overbye2020fast} creates four temporary maps and produces a cost map through weighted averaging. Both approaches combine costs in a weighted manner to generate a single-value cost map; however, they fail to consider the relationships between costs and road conditions, as well as the relationships between different costs. To address these limitations, we provide a multi-layer 2.5D feature cost map that dynamically adjusts the weight of each layer in a human-like manner according to the scenario by studying human cognition of various environmental costs under different road conditions.

%\subsubsection{Traversal Costs and Weight Allocation}
\subsection{Traversal Costs and Weight Allocation}
Planning in off-road scenarios involves considering not only the path length and smoothness but also factors like safety, efficiency, and stability. \cite{so2022sim} trained a short-horizon navigation policy using reinforcement learning (RL) without real data, which considered a goal-conditioned policy based on visual and odometry data. \cite{siva2022nauts} introduced an approach for adaptation by negotiation that enables a ground robot to adjust its navigational behaviors in uncertain, dynamic terrains. However, these methods do not consider the dynamic constraints at the trajectory planning level. Some studies, exemplified by \cite{05_park2015homotopy} and \cite{19_takemura2021traversability}, primarily focus on the constraints imposed by off-road terrains on vehicle dynamics, aiming to generate motion plans that satisfy these dynamic constraints. However, these studies do not adequately consider the environmental impact. In contrast, \cite{13_hu2021integrated} incorporates elevation information and obstacle potential fields, combining them through weighted aggregation, although the specific weight allocations are not explicitly provided. Similarly, \cite{18_weerakoon2022terp} accounts for multiple costs, including distance, orientation, obstacles, and slope. In \cite{30_li2022human}, the planning trajectory is optimized using a cost function that integrates safety, consistency, smoothness, and trajectory deviation. \cite{14_thoresen2021path} employs a hybrid A* algorithm and introduces a cost for traversability into the original objective function. \cite{castro2023does} proposed a predict costmap based on self-supervised method and used a lethal height costmap for obstacle avoidance, which weight of the two costmaps is found empirically. Nevertheless, these weight allocations are provided directly without detailed explanations for their allocation or empirical exploration to determine the weight balancing. This lack of exploration of weight allocations potentially restricts their applicability beyond experimental scenarios.

Several studies have addressed the issue of adaptability in determining cost weights for different scenarios. For instance, \cite{04_waibel2022rough} introduces a weight ratio 'e' to balance the Euclidean distance and roughness costs, highlighting the need for varying 'e' to achieve optimal trajectory planning across diverse scenarios. Similarly, \cite{16_ji2018adaptive} suggests adjusting cost weights based on user preferences, such as favoring shorter or safer (fewer pose variations) paths. However, in the context of off-road driving, the prioritization of trajectory efficiency, stability, or safety may need to be flexibly adjusted to account for changing terrains. Thus, the planning process necessitates adaptive weight adjustments to accommodate various off-road environments. While \cite{17_liu20223d} proposes using fuzzy reasoning to adaptively modify the weights of roll comfort and potential fields, this approach relies primarily on rule-based lookup using empirical rules, which often have limited applicability to specific scenarios. In contrast, the proposed method in this paper offers an adaptive weighst allocation mechanism for multiple costs, based on human driving experience, enabling effective adaptation to different scenarios.

\subsection{Human-like Research on Planning}
Human drivers possess extensive driving experience and skills, characterized by attributes such as efficiency, safety, flexibility, and stability. Several human-like algorithms aim to learn personalized driving behavior styles exhibited by human drivers. For example, \cite{26_yang2021personalized,27_yan2022driver} study the individualized driving behaviors of different drivers during lane-changing scenarios to establish personalized driving preferences and steering characteristics. \cite{30_li2022human} focuses on learning longitudinal speed planning for distinct driving styles, while \cite{39_chen2022towards} clusters driving data to identify three driving styles for speed control strategies. Additionally, \cite{40_graf2018trajectory} learns and defines two driving styles, steady and aggressive, to generate reference trajectories for planning purposes. However, these methods primarily address human driving speed habits without exploring the influence of human-like behavior on trajectory planning. To overcome this limitation, \cite{34_sadat2019jointly} proposes a joint learning approach for behavior and trajectory planning, ensuring that the generated planning trajectories exhibit human-like behavioral characteristics. Similarly, \cite{38_inoue2019robot} initially generates a large number of favorable paths using the RRT method and then employs LSTM networks to learn obstacle-avoiding path statistics for path selection, resulting in high-quality paths. Nevertheless, the human-like trajectories generated by these methods fail to adequately consider environmental factors.

Several studies have investigated human-like planning with respect to environmental factors. For instance, \cite{21_balal2016binary,22_wang2021intelligent} utilize spatial information of surrounding vehicles as input and employ fuzzy inference to acquire knowledge about a driver's trajectory planning behavior during lane-changing. Similarly, \cite{23_he2018human,24_xu2020learning,25_xu2018naturalistic} utilize natural driving data to learn from and develop human-like lane-changing planning. \cite{32_liu2022human} extracts and generalizes human driving behavior patterns when encountering obstacles, which are then applied to the planning process. \cite{yu2018human} utilized game theory to mimic human drivers' behaviors during lane-change. \cite{zhang2020multi} employed Bayesian learning methods to extract driving primitives and subsequently used a sampling-based method to generate planning trajectories. These approaches typically focus on learning specific human-like planning strategies for particular scenarios, often within simplified environments such as rule-based urban roads, while our approach can be applied more broadly in off-road scenarios with complex terrain. There are also end-to-end learning methods, \cite{33_gao2017intention} directly learn planning trajectories from general environments. However, these trajectories do not consider vehicle dynamics and cannot be readily utilized for trajectory tracking by the control system.

In this paper, we propose a human-like planning method that considers the constraints imposed by vehicle dynamics. Our approach aims to learn how humans perceive the driving environment and the patterns observed in human-driven behavior. It enables flexible adjustment of the optimal allocation of environmental costs in complex and dynamic off-road environments. Consequently, our method generates off-road spatiotemporal motion planning that best suits the vehicle's characteristics.

\section{Learning Human-like Cognition with a CNN-LSTM Network}
\label{network}
We learn human-like cognition through the process of predicting the human-like trajectories of vehicles from the environmental information processed by point clouds. We propose a CNN-LSTM network to learn the driving behavior of human drivers in different off-road scenarios. The network takes several maps as input, first extracts features with a CNN network, and then learns the trajectories in temporal order by LSTM, where the label is the trajectories with behavioral primitives attached to fit the ground truth of human driving.

\subsection{Environmental Data Extraction}
The scenario under study is a desert off-road environment with large undulations, bumpy roads, complex and variable physical properties, and few dynamic or static obstacles. Therefore, the environment is extracted in the form of a terrain elevation map, a road surface roughness map, and an artificial potential field map based on obstacle and guidance information. The extracted environment information is used as input to the network with the same dimensions and coordinate system.
\begin{figure}[h]
	\centering
	\includegraphics[width=0.45\linewidth]{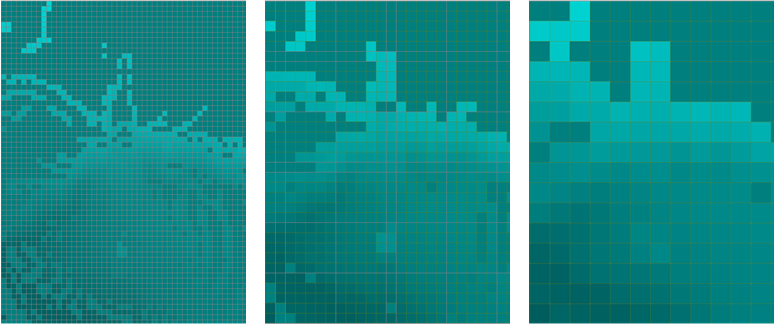}
	\caption{Three different resolutions of maps.}
	\label{data-proces}
\end{figure}

\begin{enumerate}[label=\textbf{(\alph*)}]
	\item \textbf{ The terrain elevation map ${{M}_{h}}$:}
	
Each point in the point cloud captured at the time $t$ is projected by its $x-, y-value$ onto the corresponding grid cell ${{G}_{i}}$ in the horizontal plane. Each grid cell records the $z-values$ of the points projected onto the cell and calculates the average $\mu_{z}^{G}$, mean square $\delta _{z}^{G}$, and maximum difference $\varepsilon_{z}^{G}$.

Since the point cloud is relatively sparse, there are many null values in the high resolution map, while the low resolution cannot express the environmental details. To solve this problem, three types of grid maps are generated with low, medium, and high resolutions, as shown in Fig.\ref{data-proces}. If the grid map with the highest resolution has a null cell, the value is read from the lower resolution map. The lowest resolution grid map is initialized with blank values to ensure that there are no null values. Converts the grid map with information into the terrain elevation map ${{M}_{h}}$ in pixel format. 

The value of each pixel is transformed by the z-average value $\mu _{z}^{G}$ of the corresponding grid cell.

\begin{equation}
	{{M}_{h}}(x,y)=\mu _{z}^{{{G}_{xy}}}=\frac{1}{{{n}_{xy}}}\sum\limits_{i}^{{{n}_{xy}}}{{{z}_{i}}}
\end{equation}

	\item \textbf{ The road surface roughness map ${{M}_{r}}$:}

	The road surface roughness map ${{M}_{r}}$has the same dimensions and resolution as the terrain elevation map ${{M}_{h}}$, and both maps share a similar approach for filling in missing values. 
	
	The value of each pixel ${{M}_{r}}(x,y)$ is transformed by the mean square $\delta _{z}^{G}$ of the corresponding grid cell. 
	
\begin{equation}
{{M}_{r}}(x,y)=\delta _{z}^{{{G}_{xy}}}=\frac{1}{{{n}_{xy}}}\sum\limits_{i}^{{{n}_{xy}}}{{{({{z}_{i}}-\mu _{z}^{{{G}_{xy}}})}^{2}}}
\end{equation}
	\item \textbf{ The artificial potential field map ${{M}_{f}}$:}

	\textbf{ The obstacle map ${{M}_{o}}$:} ${{M}_{o}}$ is a binary image and its generation process is similar to ${{M}_{h}}$. The pixel value "1" means that there is an obstacle and "0" means that there is no obstacle in the corresponding grid cell. If the maximum difference $\varepsilon _{z}^{G}$ in the grid cell is bigger than the threshold, i.e. $\varepsilon _{z}^{G}>{{\varepsilon }_{threshold}}$, then the grid cell is considered to have an obstacle and vice versa.
	
	\textbf{ The global reference trajectory and the guide point $p_i$:} The global reference trajectory, an outcome of global path planning, represents a route connecting the starting and ending points without factoring in obstacles or road terrain conditions. 
	
	The guide point $p_i$ denotes the intersection between the global reference trajectory and the boundary established by the vehicle's current environmental maps. Its primary function is to provide the vehicle with short-term driving direction within the immediate timeframe.
		
	\textbf{ The generation of ${{M}_{f}}$:} The artificial potential field is composed of the gravitational field formed by the guide point and the repulsive field formed by the obstacle. Where the obstacle location is provided by the obstacle map ${{M}_{o}}$. Each pixel's potential value is:
	
\begin{equation}
{{U}^{{{p}_{i}}}}=U_{att}^{{{p}_{i}}}+U_{req}^{{{p}_{i}}}
\end{equation}
$U_{att}^{{{p}_{i}}}$ is the gravitational field that guides the vehicle to the target position and $U_{req}^{{{p}_{i}}}$ is the repulsive field that steers the vehicle to avoid obstacles. 

The gravitational field is defined as:

\begin{equation}
\label{fun_4}
U_{att}^{{{p}_{i}}}=\left\{ \begin{matrix}
\frac{1}{2}\theta {{[d({{p}_{i}},{{p}_{goal}})]}^{2}}, & d({{p}_{i}},{{p}_{goal}})\le {{d}^{*}}  \\
\theta {{d}^{*}}\cdot d({{p}_{i}},{{p}_{goal}})-\frac{1}{2}\theta {{({{d}^{*}})}^{2}}, & d({{p}_{i}},{{p}_{goal}})>{{d}^{*}}  \\
\end{matrix} \right.
\end{equation}

The repulsive field is defined as:

\begin{equation}
\label{fun_5}
U_{\text{req}}^{{{p}_{i}}}=\left\{ \begin{matrix}
\frac{1}{2}\eta {{[\frac{1}{D({{p}_{i}},{{p}_{obs}})}-\frac{1}{{{D}^{*}}}]}^{2}} & D({{p}_{i}},{{p}_{obs}})\le {{D}^{*}}  \\
0 & D({{p}_{i}},{{p}_{obs}})>{{D}^{*}}  \\
\end{matrix} \right.
\end{equation}
$\theta ,\eta $ are the fitness factors, can be used to adjust the value range. ${p}_{goal}$ represents the coordinates of the target point, which refers to the endpoint of the reference line within the current map range. ${p}_{obs}$ represents the coordinates of the obstacle. $d(,)$ and $d^*$ represent the distance between two points and calculation boundary in the gravitational field. $D(,)$ and $D^*$ represent similarly in the gravitational field.
	\item  \textbf{ The ego momentum map ${{M}_{u}}$:}

The ego state information $u=({{v}_{x}},{{v}_{y}})$ is extracted from the inertial measurement unit(IMU) data. For uniformity inputs, the momentum map ${{M}_{u}}$ is in the same coordinate system, size and resolution as the four maps in a)-d). 

The momentum map satisfies the following 3 assumptions: i) the further away from the ego, the lower the momentum; ii) the larger the angle with the direction of the motion, the lower the momentum; iii) the higher the speed, the higher the momentum. Therefore, the momentum of each pixel on the map satisfies the following equation:
\begin{equation}
\label{fun_6}
M_u(x,y) = \zeta \left| {({v_x},{v_y})} \right| \cdot \frac{{(x,y) \cdot ({v_x},{v_y})}}{{\left| {(x,y)} \right|\left| {({v_x},{v_y})} \right|}} \cdot \left( {1 - \frac{{\left| {(x,y)} \right|}}{{\left| {({v_x},{v_y})} \right|}}} \right)
\end{equation}
Simplify as: 
\begin{equation}
M_u(x,y)=\zeta \left( x\cdot {{v}_{x}}+y\cdot {{v}_{y}} \right)\left( \frac{1}{\left| (x,y) \right|}-\frac{1}{\left| ({{v}_{x}},{{v}_{y}}) \right|} \right)
\end{equation}
$M_u(x, y)$ represent the momentum value of the pixel in map. $\zeta $ is the fitness factor, can be used to adjust the value range.
	
\end{enumerate}

\subsection{CNN-LSTM Network}
\begin{figure}[h]
	\centering
	\includegraphics[width=7.2in]{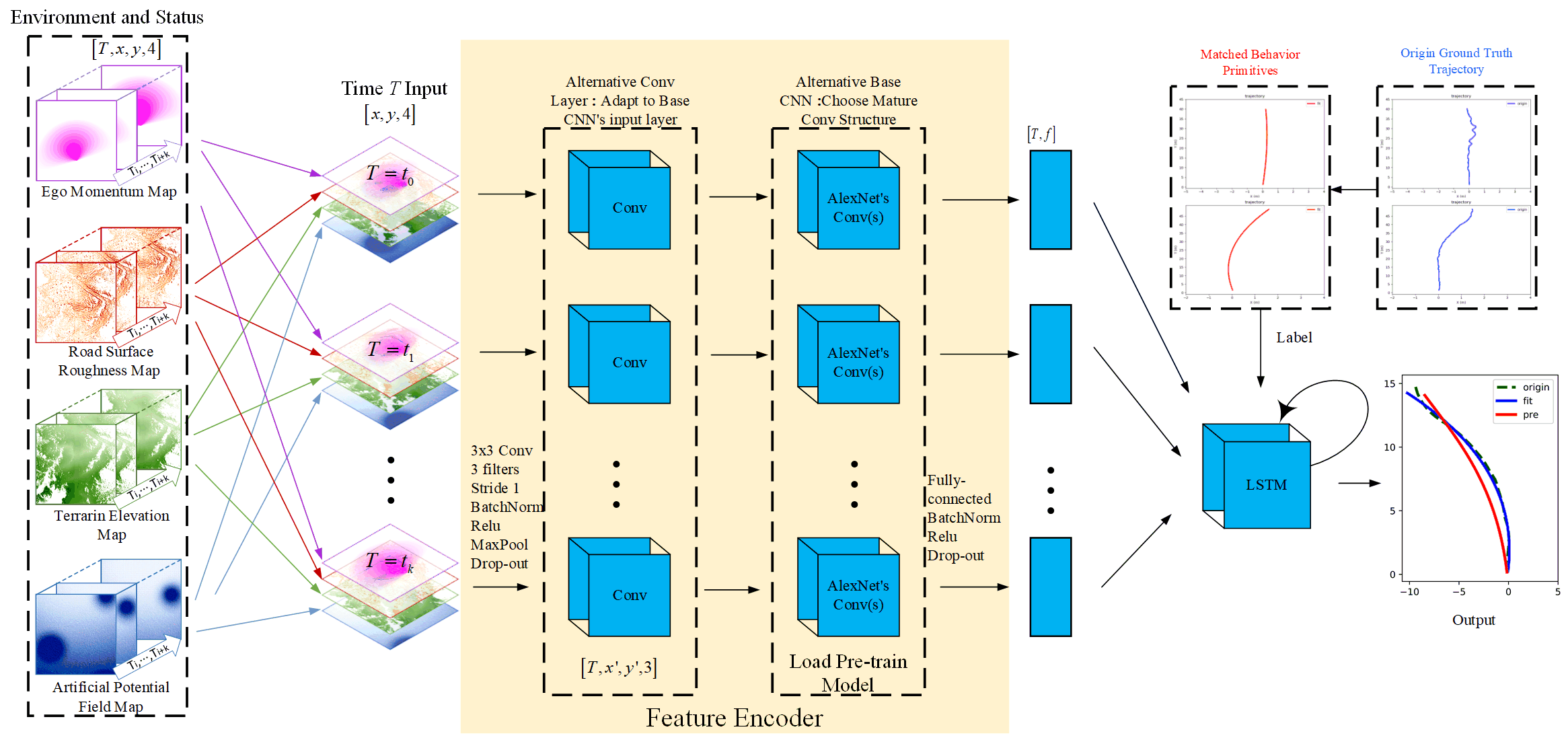}
	\caption{The CNN-LSTM network learns about sequences of human driving trajectories. On the left, a sequence of environmental maps is input into the network sequentially over $T$ consecutive time steps. A CNN module is employed as the feature encoder, wrapping map information at each time step. The output is then fed into the LSTM module on the right, with the network's label representing the ground truth trajectory.}
	\label{cnn-lstm}
\end{figure}

To predict the human-like planning trajectory of the vehicle based on the input environmental image information, we employed a CNN-LSTM network, as illustrated in Fig.\ref{cnn-lstm}. A sequence of environmental maps, including terrain elevation, road surface roughness, artificial potential field, and ego momentum maps, is sequentially input into the network over $T$ consecutive time steps.

\begin{enumerate}[label=\textbf{(\alph*)}]
	\item \textbf{Feature encoder:} 

To extract features from the environmental maps, a CNN module is employed as the feature encoder to wrap map information at each time step. The selection of a well-established backbone network is flexible in order to mitigate training expenses. We choose the AlexNet architecture as the backbone network.
 \item \textbf{Sequence prediction:}

We feed encoded environmental features into the LSTM module to acquire knowledge about sequences of human driving trajectories. The resultant trajectory sequence generated by the network has a length of $2H$ and comprises of $x$ and $y$ positions, sharing the same length as the ground truth trajectory.
	\item \textbf{Ground truth and labels:}

The ground truth of the trajectory is extracted from the GPS and IMU data. First, we convert the GPS data to UTM coordinates, then we use the start point as the coordinate origin and the start heading as the y-axis direction. The vehicle coordinate values of the future discrete time $\tau =\{{{\tau }_{1}},{{\tau }_{2}},\cdots ,{{\tau }_{H}}\}$ is calculated as the ground truth and noted as $\Gamma _{gt}^{\tau }$.

Due to the jitters and instabilities arising from vehicle movements, the trajectory ground truth obtained from GPS is unsuitable for direct usage as labels in the network. Therefore, we match the ground truth trajectory ${{\Gamma }_{gt}}$ with the ego's behavioral trajectory clusters (as described in the following Section \ref{back}), with the closest matching trajectory deemed as the fitted trajectory ${{\tilde{\Gamma }}_{gt}}$ and utilized as the label for the network.
	\item \textbf{Loss function:}

The loss function of the network is designed to measure the distance between the output ${{\tilde{\Gamma }}_{pre}}$ and the ground truth trajectory ${{\Gamma }_{gt}}$, and is formulated as follows:

\begin{equation}
d\left( {{{\tilde{\Gamma }}}_{pre}},{{{\tilde{\Gamma }}}_{gt}} \right)=\frac{1}{H}\sum\limits_{h=1}^{H}{{{\left\| \tilde{\Gamma }_{pre}^{h}-\tilde{\Gamma }_{gt}^{h} \right\|}_{2}}}
\end{equation}
\end{enumerate}

\section{Motion planning guided by human-like cognition}
\label{planner}
\begin{figure}[h]
	\centering
	\includegraphics[width=6.8in]{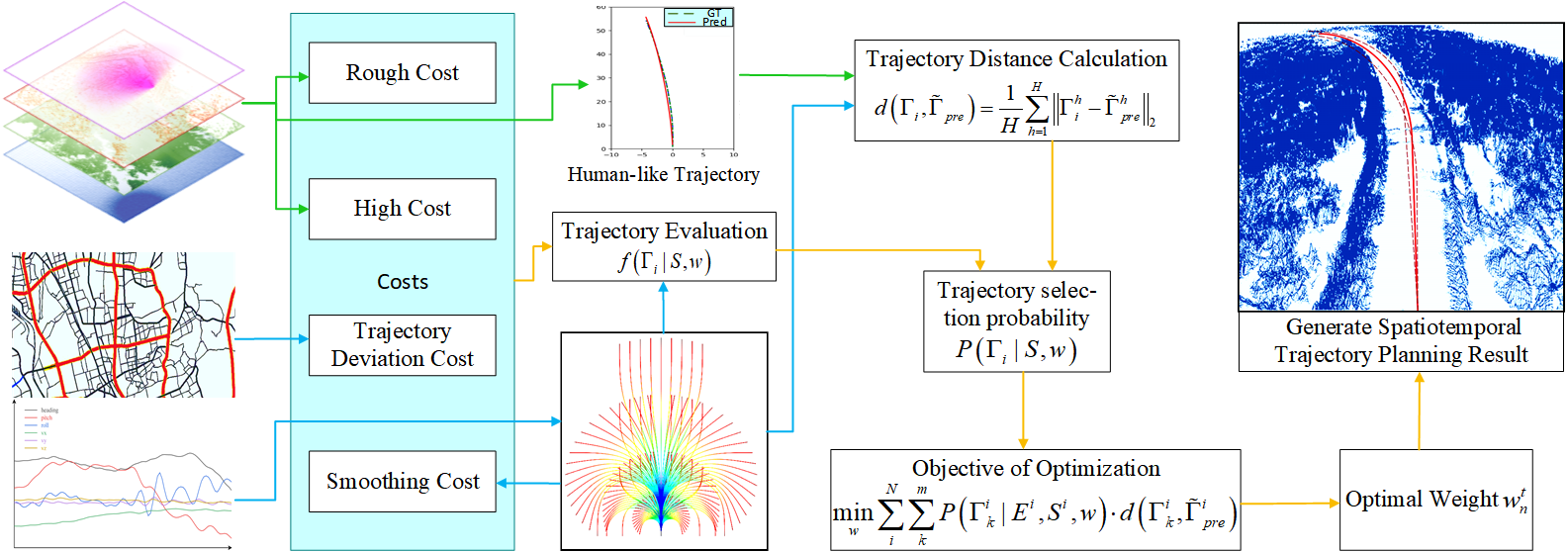}
	\caption{Motion planning guided by human-like cognition. The green line represents the process of obtaining human-like trajectories based on environmental inputs. The blue line represents the process of generating clusters of primitive actions based on the ego's state. The yellow line represents the process of solving the cost weights associated with human-like behaviors and ultimately generating the spatiotemporal motion planning results.}
	\label{human-like}
\end{figure}

The complete workflow for motion planning guided by human-like cognition is shown in Fig.\ref{human-like}. We first learn the best cost weights in current scenarios based on human-like trajectories, then select and extend the motion primitives based on the cost weights, and generate the spatiotemporal motion planning results.
\subsection{Cost in Planning}
In this environment, the four components of the basic cost $J$ of planning are: terrain height cost ${{J}_{h}}$, road roughness cost ${{J}_{r}}$, trajectory deviation cost ${{J}_{\Gamma }}$, and smoothness cost ${{J}_{s}}$.

\begin{enumerate}[label=\textbf{(\alph*)}]
	\item \textbf{ Terrain height cost:}

The height of each point on the trajectory are provided by the terrain elevation map ${{M}_{h}}$. The height value itself is not significant. Therefore, the first derivative of the trajectory point is defined as the height cost $C_{h}^{k}$ at that point. We use the Central Finite Difference method to provide the height cost for each point on the trajectory, as shown in Equ.(\ref{CFD}).
\begin{equation}
\label{CFD}
C_{h}^{k}=\frac{\mu _{z}^{{{G}_{k+1}}}-\mu _{z}^{{{G}_{k-1}}}}{\left| {{G}_{k+1}}-{{G}_{k-1}} \right|}
\end{equation}
$\left| {{G}_{k+1}}-{{G}_{k-1}} \right|$ is the distance between point $k+1$ and point $k-1$.

The mean of the height cost of all points on the trajectory is defined as the height cost ${{J}_{h}}$ of the trajectory, as shown in Equ.(\ref{h-cost}).

\begin{equation}
\label{h-cost}
{{J}_{h}}=\frac{\sum\limits_{i}^{H}{C_{h}^{i}}}{H}
\end{equation}
	\item \textbf{Road roughness cost:}

The roughness value of each point on the trajectory is provided by the road surface roughness map ${{M}_{r}}$. The mean roughness value of all points on the trajectory is defined as the roughness cost ${{J}_{r}}$ of the trajectory, as shown in Equ.(\ref{R-cost}).
\begin{equation}
\label{R-cost}
{{J}_{r}}=\frac{\sum\limits_{i}^{H}{\delta _{z}^{{{G}_{i}}}}}{H}
\end{equation}
	
	\item \textbf{Trajectory deviation cost:}

The reference trajectory is the input of the motion planning trajectory system, providing the most basic reference and guidance. We use Gaussian functions as weight coefficients for the deviation of trajectory points. The cost ${{J}_{\Gamma }}$ of the deviation between primitives and the expected trajectory is shown as Equ.(\ref{t-cost}):
\begin{equation}
\label{t-cost}
{{J}_{\Gamma }}=\frac{\sum\limits_{n=0}^{{{N}_{p}}}{g\left( 1-\frac{n}{{{N}_{p}}} \right){{j}_{n}}}}{{{N}_{p}}+1}
\end{equation}
${{N}_{p}}$ represents the number of equidistant trajectory points taken into account for the deviation, $g\left( \cdot  \right)$ is the standard normal distribution of the Gaussian function, and ${{j}_{n}}$ represents the cost of the deviation between the generated trajectory and the reference trajectory.

\begin{equation}
\label{fun_13}
{{j}_{n}}={{w}_{d}}{{d}_{n}}+{{w}_{\theta }}\Delta {{\theta }_{n}}
\end{equation}
${{d}_{n}}$ and ${{w}_{d}}$ represent the distance deviation and its weight coefficient, while $\Delta {{\theta }_{n}}$ and ${{w}_{\theta }}$ represent the heading deviation and its weight coefficient.
	
	\item \textbf{Smoothness cost:}

The smoothness of the trajectory reflects the average change of the curvature of the primitives at each point, and the smaller the value, the smoother the curvature. The smoothness of the trajectory is defined as shown in Equ.(\ref{S-cost}).
\begin{equation}
\label{S-cost}
{{J}_{s}}=\sum\limits_{n=2}^{{{N}_{P}}}{\frac{\left( \kappa _{n-1}^{2}+\kappa _{n}^{2} \right)\Delta {{s}_{n}}}{2}}
\end{equation}
${{\kappa }_{n}}$ is the curvature value of the primitive sampling point, and $\Delta {{s}_{n}}$ is the distance between adjacent sampling points.

\end{enumerate}

\subsection{Human-like Cost Weights Learning}
\label{weight}
The motion planning algorithm typically selects an optimal trajectory ${{\Gamma }_{k}}$ from a set of trajectories (The generation of trajectories in our paper is described in detail in Section \ref{back}) generated within the vehicle state $S$ and environment $E$. 
\begin{equation}
\arg {{\min }_{k}}f\left( {{\Gamma }_{k}}|S,E,w \right)
\end{equation}
$f$ represents the cost function, where its parameters $w$ denote a set of weights utilized to balance various costs based on their respective importance. The cost function $f$ of the trajectory in our context is:

\begin{equation}
f\left( {{\Gamma }_{k}}|S,E,w \right)={{w}_{h}}{{J}_{h}}+{{w}_{r}}{{J}_{r}}+{{w}_{\Gamma }}{{J}_{\Gamma }}+{{w}_{s}}{{J}_{s}}
\end{equation}
${{w}_{h}},{{w}_{r}},{{w}_{\Gamma }},{{w}_{s}}$ represent the weighting coefficients of the corresponding cost indicators. The weight set $w$ denotes the weights of all costs:

\begin{equation}
w=({{w}_{h}},{{w}_{r}},{{w}_{\Gamma }},{{w}_{s}})
\end{equation}

In traditional methods, these cost weights in $w$ are manually adjusted as a fixed set of parameters. Consequently, the planned trajectories selected in traditional planners may neither align with human-like driving behaviors nor adapt to the current environment.
	
Therefore, in this subsection, we introduce a method for learning optimal cost weights in present scenarios. This method utilizes human-like trajectories generated by the network trained in Section \ref{network}. The yellow line in Fig.\ref{human-like} illustrates the solving process.

To learn the optimal weights, we construct a two-layer optimization objective to translate this into an optimization problem. Before proceeding, we define the distance metric, denoted as $d({{\Gamma }_{k}},{{\tilde{\Gamma }}_{pre}})$, which calculate the distance loss between each trajectory primitive and predicted human-like trajectory as follows:
\begin{equation}
d\left( {{\Gamma _k},{{\tilde \Gamma }_{pre}}} \right) = \frac{1}{H}\sum\limits_{h = 1}^H {{{\left\| {\Gamma _k^h - \tilde \Gamma _{pre}^h} \right\|}_2}}
\end{equation}
Subsequently, we construct the first layer of the optimization objective to calculate the total distance loss. One straightforward objective function we can use is the Mixture-of-Experts (ME) loss \cite{jacobs1991MEloss}, defined as:
\begin{equation}
L_{i}^{ME}=\sum\limits_{k}^{m}{P\left( {{\Gamma }_{k}}|E,S,w \right)}\cdot d\left( {{\Gamma }_{k}},{{{\tilde{\Gamma }}}_{pre}} \right)
\end{equation}
$m$ represents the number of behavioral trajectory clusters. The ME loss is one of the common losses in multimodal deep learning. We adopt it to assign a higher selection probability to trajectory primitives that closely resemble predicted human-like behavior. $P\left( \cdot  \right)$ is the probability of the selection of the trajectory ${\Gamma _k}$, a softmax transformation for mapping a cost to a probability:
\begin{equation}
P\left( {{\Gamma }_{k}}|E,S,w \right)=\frac{{{e}^{-f({{\Gamma }_{k}}|E,S,w)}}}{\sum\nolimits_{j}^{m}{{{e}^{-f({{\Gamma }_{j}}|E,S,w)}}}}
\end{equation}

Single-time solutions might exhibit randomness or overfitting. So, we establish the second layer of the optimization function where we conduct optimizations over a short, consecutive frame time:
\begin{equation}
\underset{w}{\mathop{\min }}\,\sum\limits_{i}^{N}{L_{i}^{ME}}
\end{equation}
$N$ denotes the number of consecutive optimizations performed over a period of time. Hence, the complete two-layer objective function is:
\begin{equation}
\label{min}
\underset{w}{\mathop{\min }}\,\sum\limits_{i}^{N}{\sum\limits_{k}^{m}{P\left( {{\Gamma }_{k}}|E,S,w \right)}\cdot d\left( {{\Gamma }_{k}},{{{\tilde{\Gamma }}}_{pre}} \right)}
\end{equation}

We employ the L-BFGS algorithm to solve this nonlinear optimization problem.

\subsection{Motion Planning with Behavioral Primitives}
\label{back}
\subsubsection{Generation of Behavioral Primitive Library}
\label{BPL}
Behavior categories extraction \cite{wang2020motion_23} and multiple trajectory modeling \cite{yaomin2021representation_24} have been completed in previous research work. The proposed behavioral primitive model is applicable to heterogeneous vehicles, encompassing both wheeled and tracked vehicle platforms. Further details on the constraints, including motion differential constraints and driving behavior constraints, can be found in our previous work \cite{primitives2023}. Building upon this foundation, data-driven driving behavior results are used as constraints for driving behaviors with spatiotemporal coupling information. The problem of generating primitives is transformed into an optimization problem. The definition of the optimization problem is as follows:
\begin{equation}
	\begin{array}{l}
		\mathop {\min }\limits_u :{\ }{g_{w/t}}\left( {s,u} \right)\\
		s.t.{\ \ \ \ }\mathop s\limits^ \cdot   = {f_g}\left( {s\left( t \right),u\left( t \right)} \right){\qquad}t \in \left[ {{t_1},{t_g}} \right]\\
		{\qquad \ \ \ }s(t) = {{\beta}_e}{\qquad}t = {t_1}, \cdots ,{t_1} + {N_s} \cdot \Delta {t_b}\\
		{\qquad \ \ \ }\left( {s\left( t \right),u\left( t \right)} \right) \in U{\qquad}t \in \left[ {{t_1},{t_g}} \right]\\
		{\qquad \ \ \ }\left( {s\left( t \right),u\left( t \right)} \right) = T{\qquad}t = {t_1},{t_g}
	\end{array}
\end{equation}
Here, $u$ and $s$ are respectively the control and state parameters of the vehicle. The objective function $g_{w/t}$ considers the smoothness of the trajectory. The optimization problem constraints include the motion differential constraints $f_g$ with spatiotemporal information, the driving behavior constraints ${\beta}_e$, the inequality constraints of the state and control variable values $U$, and the smooth transition constraints between primitives $T$.

The offline optimization and generation of the behavior primitive library are solved using IPOPT and CVODES solvers. The resulting vehicle behavior primitive library generated under vehicle state is denoted as: 
\begin{equation}
	B\left( S \right)={{\left\{ {{B}_{1}},{{B}_{2}},\cdots ,{{B}_{k}},\cdots  \right\}}_{\text{bn}}}
\end{equation}
${B_{k}}$ is the one behavioral primitive in the library.
\subsubsection{Generation of Behavioral Trajectory Clusters}
In our study, it is necessary to obtain behavioral trajectory clusters with arbitrary characteristics. However, the starting point of behavior primitives is subject to certain constraints. To tackle this problem, we leverage the behavior primitives library to generate new trajectory clusters that meet our desired criteria.

We first concatenate the primitives, i.e. $B({{S}_{i}})+B({{S}_{i+1}})$. Rotations and translations are applied at the junctions of primitives to ensure smoothness. We then extract trajectory segments starting from any moment of $B({{S}_{i}})$, with a duration equivalent to the predicted vehicle trajectory, to generate behavior trajectory clusters with no starting point constraint under situation ${{S}_{i}}$. The generated behavioral trajectory clusters are denoted as:
\begin{equation}
	\Gamma \left( {{S}_{i}} \right)={{\left\{ {{\Gamma }_{1}},{{\Gamma }_{2}},\cdots ,{{\Gamma }_{k}},\cdots  \right\}}_{m}}
\end{equation}
$m$ represents the number of behavioral trajectory clusters, and ${{\Gamma }_{k}}$ is the one behavioral trajectory in the cluster.

\subsubsection{Motion Planning}
The optimal selection of primitives is a comprehensive evaluation of the defined cost, and the total cost of each primitive is shown in Eq.(\ref{motion_P}).
\begin{equation}
	\label{motion_P}
	J(B_k)={{w}_{h}}{{J}_{h}}(B_k)+{{w}_{r}}{{J}_{r}}(B_k)+{{w}_{\Gamma }}{{J}_{\Gamma }}(B_k)+{{w}_{s}}{{J}_{s}}(B_k)
\end{equation}
where the $w$ is the current environment optimal weight obtained in Section \ref{weight}. 

After calculating the selection cost of the alternative primitives obtained, the primitive with the lowest cost in the primitive library is selected as the optimal expansion primitive. The primitive sequence is gradually expanded and generated from the start point to the end point, and the sequence of timed points for the planned trajectory is finally generated.

The human-like trajectory planner we proposed can acquire knowledge about how human drivers perceive the environment by analyzing predicted human-like planned trajectories, extracting the current optimal cost weights, and generating highly adaptive and human-like trajectory planning results for off-road environments. 

\section{Result}
In this section, we present the representative outcomes of proposed method from our experiments. Firstly, We intruduce the experimental setup in Section.\ref{51}.Subsequently, we present the human-like trajectory output results of the CNN-LSTM network in Section.\ref{52}.  Finally, we demonstrate the on-road experiment in Section.\ref{53} which aims to validate our human-like motion planner in providing optimal cost weights for the current situation in various scenarios.
\begin{figure}[h]
	\centering
	\includegraphics[width=0.5\linewidth]{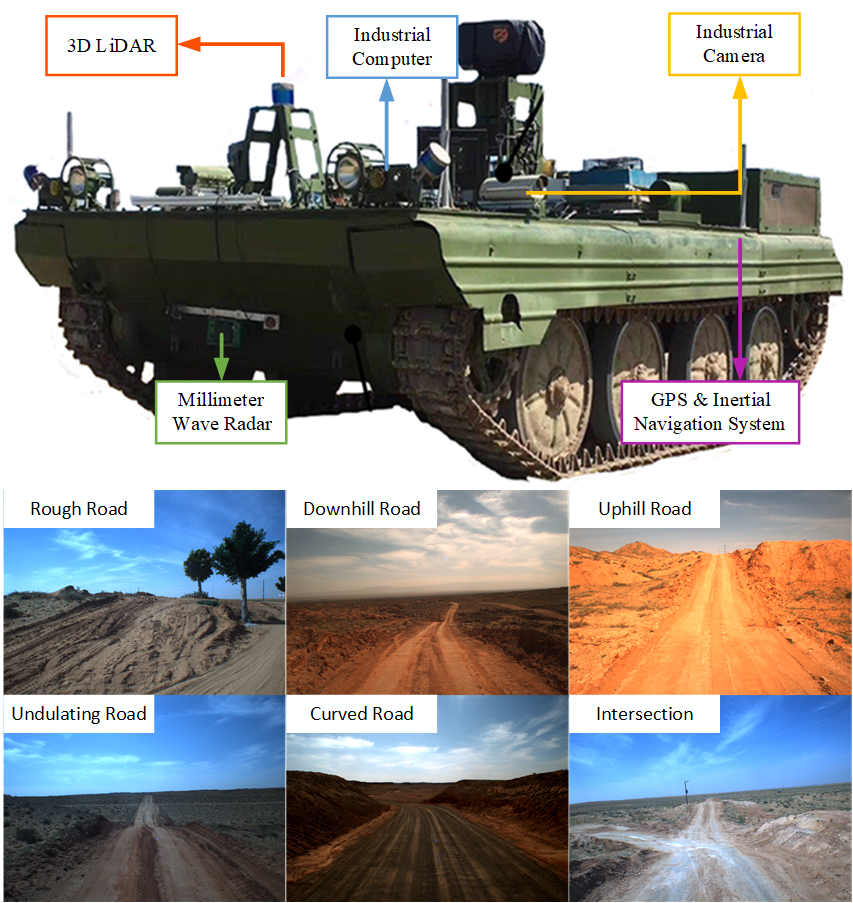}
	\caption{The vehicle platform and various testing scenarios.}
	\label{vehicle}
\end{figure}
\subsection{Experimental Setup}
\label{51}
\subsubsection{Experimental Vehicle Platform with a Sensor Suite}\label{Sensor}

The experimental vehicle platform used for data collection and validation is a dual-mode tracked vehicle, capable of both manned and unmanned driving. It is worth reiterating that the proposed behavioral primitive model is applicable to heterogeneous vehicles, encompassing both wheeled and tracked vehicle platforms.

It is equipped with a variety of sensors including Industrial Camera, 3D LiDAR, GPS, Inertial Navigation System, Industrial Computer and others as required. The vehicle platform and testing environment are depicted in Fig.\ref{vehicle}.

\subsubsection{Data Collection and Testing Configuration}
\begin{enumerate}[label=\textbf{(\alph*)}]
\item \textbf{Data for human-like trajectory prediction}

In order to train our CNN-LSTM network, we collected data recordings of driving in rugged desert off-road scenarios for a duration of 5 hours using the vehicle platform and the mounted sensors mentioned in Section \ref{Sensor}. We utilized IMU and GPS to acquire vehicle trajectory information, while dense point cloud data was obtained using a 64-line LiDAR.

The collected trajectory data have been converted to the vehicle coordinate system (with the vehicle heading direction as the y-axis, following the right-hand coordinate system principle) based on the IMU vehicle state data before use. The vehicle trajectory data has been segmented into groups every 6 seconds. This sets the predicted trajectory length $H$ for the training network to 60. For the environment map, we've chosen $\theta =0.2$ in Eq.(\ref{fun_4}), $\eta=50$ in Eq.(\ref{fun_5}), and $\zeta=42$ in Eq.(\ref{fun_6}). These parameters are adjustable according to actual environmental conditions and map dimensions, ensuring a balanced distribution across the range while preventing excessive discretization beyond the pixel representation range of the map. For cost in planning, we've set $w_d=0.8$ and $w_\theta=0.2$ in Eq.(\ref{fun_13}) .

To achieve effective training, we carefully selected the highest-quality data frames from the dataset, resulting in a final selection of approximately 32,000 frames. These frames were uniformly sampled in terms of time intervals, with 5\% of them being designated as validation frames. Validation frames were selected by excluding them from the training frames. Ultimately, we utilized 30,273 training frames and 1,593 validation frames. 

\item \textbf{On-road testing configuration of our human-like motion planner.}

During our on-road validation, we selected several representative and challenging off-road scenarios. We recorded relevant 'rostopics' of the planned results real-time into 'rosbag'. We recorded the 'rostopics' of the real-time perception multi-layer maps, which are used by other planners to generate compared trajectories offline.

Because of the rough road surfaces in various scenarios, we restricted the maximum value of the global speed to not exceed 30 km/h during the experiments. The local velocity is determined by the planning algorithm. The perception program operates at an efficiency of 10 Hz. The planning results are continually updated at the same frequency, consistently utilizing the most recent cost weights.

In the entirety of the algorithm, the neural network requires 108ms, optimization for optimal weights demands 37ms, and the planning algorithm consumes 24ms.  The optimization algorithm for cost weights operates as a separate thread, with a slightly longer runtime (totaling 145ms), which does not affect the planning algorithm (only 24ms, always use the latest cost weights) update results at a frequency of 10Hz.
\end{enumerate}

\subsubsection{Experimental Objectives and Methods}
\begin{enumerate}[label=\textbf{(\alph*)}]
\item \textbf{Experiment of human-like trajectory prediction}

The experiment of predicting human-like trajectories aims to validate the high accuracy achieved by the CNN-LSTM network proposed in Chapter \ref{network}. This network's suitability for deriving human-like planning cost weights in Chapter \ref{planner} will be assessed, leading to the development of a human-like planner.

In Section \ref{521}, we conducted ablation experiments to demonstrate the significant influence and significance of the four-layer maps used as inputs in our proposed method on trajectory prediction.

Subsequently, in Section \ref{522}, we presented the test results and compared them with a pure LSTM algorithm that does not consider environmental inputs, highlighting the superior learning effectiveness of our method in complex environments for humanoid trajectory generation.

Finally, in Section \ref{523}, we showcased and analyzed the test results of specific scenarios to discuss the factors contributing to the high prediction accuracy of our method in off-road trajectory planning.

\item \textbf{Experiment of our human-like motion planner}

The on-road experiment aims to validate our human-like motion planner proposed in Chapter \ref{planner} in providing optimal cost weights for the current situation in various scenarios. Our planner ensures that autonomous vehicles can consistently maintain human-like decision-making strategies and make the most suitable choices in current situation. 

Section \ref{531}-\ref{533} is dedicated to the detailed analysis and evaluation of our proposed method, based on the planning results of individual frames obtained during the experimental process. In different scenarios, the desired performance of planned trajectories can vary. For instance, in terrain-dominated scenarios, our ideal trajectory would have the lowest roughness. Therefore, these three sections conducted three pairs of comparative experiments to validate the adaptability of our human-like motion planner to the environment, as well as the efficiency, stability, safety, and consideration of vehicle dynamics in the planned trajectories.

Section \ref{534} presents a tabular representation of the testing outcomes at various locations along the entire test process for the aforementioned six scenarios. It should be noted that the cost weights of the planner for the current scenario in the results are continuously updated based on environmental changes. In contrast, the cost weights of the compared planner do not update dynamically, as it is limited to selecting fixed values from other scenarios. The performance of the most noteworthy planning trajectories in each scenario is visually displayed and compared through figures.

\end{enumerate}
\begin{figure}[t]
	\centering
	\includegraphics[width=0.65\linewidth]{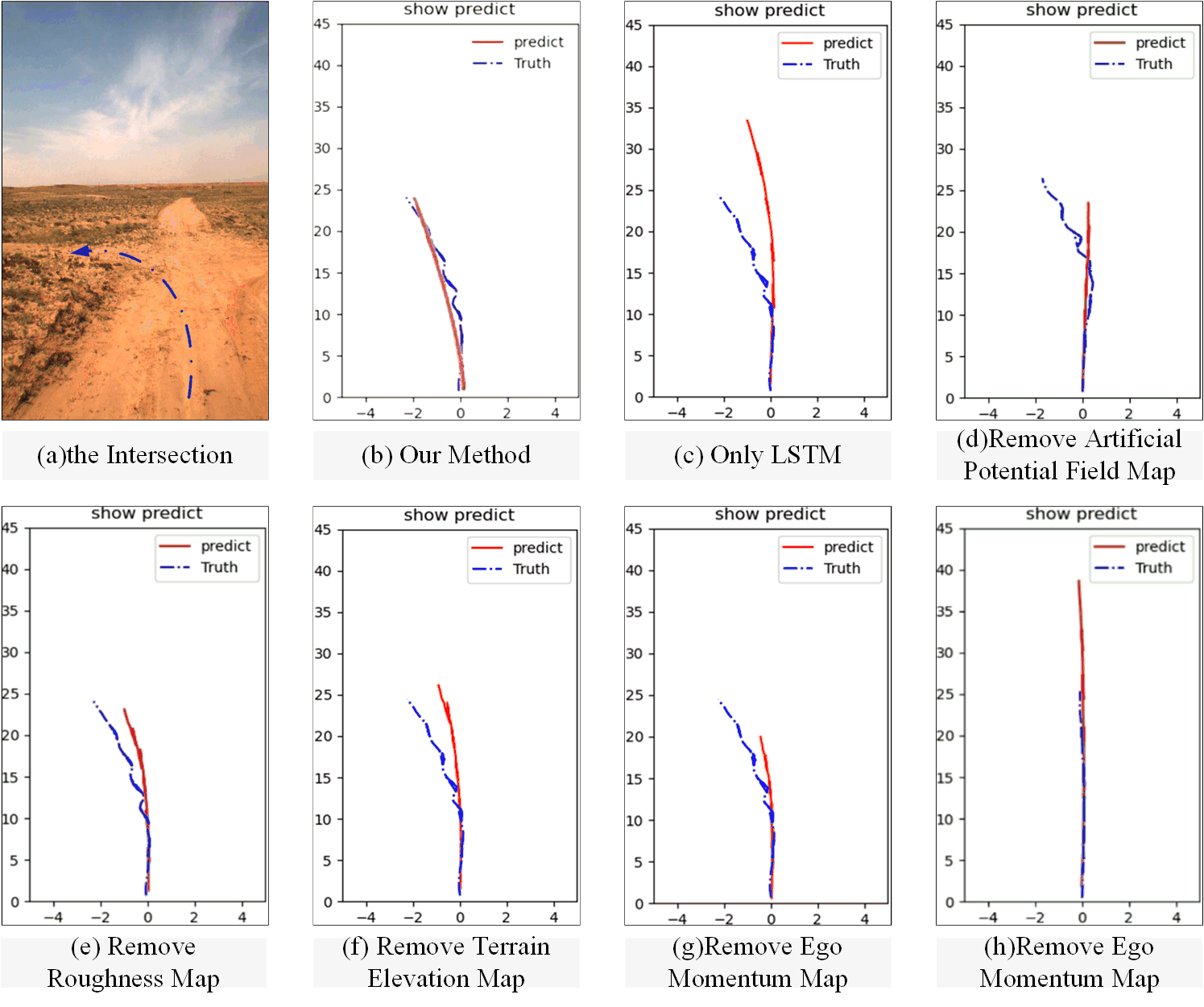}
	\caption{Ablation Experiment Results at a Poorly Defined Intersection. The subfigures display the outcomes when each map is individually removed.}
	\label{ablation}
\end{figure}
\subsection{Experiment of Human-like Trajectory Prediction}
\label{52}
The experiment of predicting human-like trajectories aims to validate the high accuracy achieved by the CNN-LSTM network proposed in Chapter \ref{network}. The experimental process begins with ablation experiments in Section \ref{521}, followed by presenting the test results in Section \ref{522}, and concluding with an analysis and discussion of the findings in Section \ref{523}. The experimental outcomes confirm the suitability of our human-like trajectory predictions to meet the requirements.
\subsubsection{The Contribution of Each Map}
\label{521}

Table \ref{ADE} presents the results of an ablation study in which we sequentially remove one map from the input set. It is evident that when the terrain elevation map or road surface roughness map is removed, there is a significant increase in the average displacement error (ADE) and the final displacement error (FDE) errors . When the artificial potential field map is omitted, the errors exhibit a slight increase, with a notably higher increase in FDE. Removal of the ego momentum map results in a notable increase in both average error and a slight rise in FDE.
\begin{table}[h]
	\caption{The ADE and FDE of the prediction results of an ablation study}
	\label{ADE}
	\centering
	\begin{tabular}{c|cc}
		\hline
		Condition & ADE   & FDE   \\ \hline
		Remove Terrain elevation map & 0.457 & 0.671 \\
		Remove road surface roughness map & 0.512 & 0.799 \\
		Remove artificial potential field map & 0.279 & 0.814 \\
		Remove ego momentum map & 0.592 & 0.422 \\
		Only LSTM        & 0.724 & 1.097 \\
		Our Methods & 0.136 & 0.371 \\ \hline
	\end{tabular}
\end{table}

We use a prediction at a poorly defined intersection as an example to discuss the ablation experiments. Fig.\ref{ablation}(b) represents the results of our method in which the predicted path closely aligns with ground truth. In contrast, Fig.\ref{ablation}(c) shows the predictions from a pure LSTM model, which only infers a leftward trend but produces a path that does not conform to the road conditions. 

Fig.\ref{ablation}(d), depicting the absence of the potential field map, resembles input data without guiding point information and consequently fails to predict a turn, providing predictions for straight-ahead movement instead. Fig.\ref{ablation}(e) and (f), illustrating the absence of the elevation and roughness maps, respectively, both exhibit inaccurate detection of right-angle turns, leading to less precise predictions of turning trajectories. Fig.\ref{ablation}(g), depicting predictions without the momentum map, shows a generally correct trend in predicted trajectories, but with noticeably slower speeds than actual speeds. Moreover, predictions without the momentum map exhibit significant issues when the vehicle's speed changes, as shown in Fig.\ref{ablation}(h).

These findings indicate that all four input maps in our proposed algorithm contribute to the improvement of prediction accuracy. Combining all four maps yields results that are more favorable for prediction outcomes.

\subsubsection{Result}
\label{522}

We use the evaluation indicators based on trajectory prediction to evaluate the predicted results of human-like motion trajectories. The ADE is 0.136, and the FDE is 0.371.

Due to a lack of algorithms capable of directly predicting the trajectory of the ego vehicle in a manner that emulates human driving behavior, there is no basis for comparison. Therefore, the work of this subtask is only compared to the LSTM method, which predicts based on the historical trajectory. As shown in Table \ref{ADE}, the ADE and FDE of the LSTM trajectory prediction results are 0.724 and 1.097, respectively. It can be observed that our proposed method performs significantly better than the LSTM method in predicting the trajectory of the ego vehicle.

\subsubsection{Discussion}
\label{523}
\begin{figure}[h]
	\centering
	\includegraphics[width=0.55\linewidth]{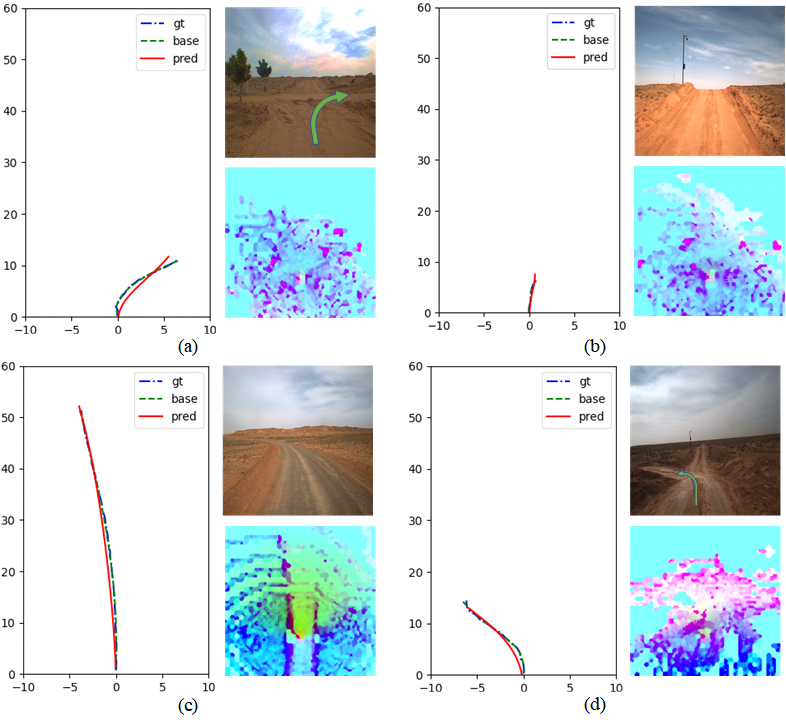}
	\caption{It shows the predicted results of the human-like trajectory. The left side of each sub-figure shows the ground truth trajectory (the blue line "gt"), the ground truth fitting (the green line "base"), and the predicted result (the red line "pred"). The upper right side shows the scenario, while the lower right side presents the fused result of the environment map(visualized by superimposing it on the RGB layer) at the prediction time. Specifically, (a) presents the right turn scenario, (b) presents the uphill scenario, (c) presents the long curve scenario, (d) presents a left forward steering scenario.}
	\label{Pred}
\end{figure}
The detection results of some scenarios in the test dataset are shown in Fig.\ref{Pred}. As shown, our human-like trajectory generation method performs well in various scenarios, including turns, long curves, and those that require acceleration or deceleration. The results obtained by our method is more accurate than other methods used to predict the trajectories of other vehicles.
\begin{enumerate}[label=\textbf{(\alph*)}]
	\item{\textbf{Accurate speed prediction:} We selectively extract terrain-related information from environmental data and feed it into the network to help us accurately predict speed. In Fig.\ref{Pred}(a),(c) and (d), our method takes into consideration the roughness of road surface from the  road surface roughness map ${{M}_{r}}$.Thus our method predicts a low speed in the rough scenario in (a) and (d), and a high speed in smooth surface in Fig.\ref{Pred}(c), which are validated by ground truth. Fig.\ref{Pred}(b) illustrates an uphill scenario where the predicted trajectory also accurately predicts the speed reduction due to the height information extracted from the terrain elevation map ${{M}_{h}}$.}
	
	\item{\textbf{Matching shapes:} Our network input is created by layering multiple environmental features at the same size and resolution, thereby ensuring comprehensive information coverage per unit area. It makes our trajectory almost match the shapes of the ground truth. The predicted trajectory almost coincides with the ground truth, especially in Fig.\ref{Pred}(c), the long-curve scenario is a challenge for planning, which need the trajectory fit the shape of the road as closely as possible. It demonstrates that layers of terrain feature input play a critical role in accurately predicting vehicle trajectories.}
	
	\item{\textbf{Correct trajectory selection:} The input for ego vehicle trajectory prediction includes guide points and global planning trajectory information, that allows the correct trajectory to be selected. Fig.\ref{Pred}(a) shows our accurate prediction of a right-turn behavior by using the guidance information obtained from an artificial potential field map${{M}_{f}}$. Furthmore, in the case of a three-way intersection shown in Fig.\ref{Pred}(d), the predicted trajectory accurately chooses the same trajectory as the ground truth, also attributing to the method of using guide information as network input. Due to the lack of strong constraints on the mechanical properties of the vehicle, the directly output trajectories do not correspond to the vehicle motion model and cannot be provided directly to the vehicle controller. So we use the predict trajectories to solve the optimal cost-weight of planning. We show our planning result base on the optimal weight. Our trajectory prediction meets the requirement of providing a human-like cognitive benchmark for solving the optimal cost-weight of planning.}
\end{enumerate}

\subsection{Experiment of Our Human-like Motion Planner}
\label{53}
The on-road experiment aims to validate our human-like motion planner proposed in Chapter \ref{planner} in providing optimal cost weights for the current situation in various scenarios. Our planner ensures that autonomous vehicles can consistently maintain human-like decision-making strategies and make the most suitable choices in current situation. The experimental results of keyframes for three pairs of scenarios are in Section \ref{531}-\ref{533}, as well as the statistical data of the overall testing performance for each scenario in Section \ref{534}.

This section validated the adaptability of our human-like motion planner to the environment and the efficiency, stability, safety, and consideration of vehicle dynamics in the planned trajectories. The optimal cost weight results for the six scenarios, including the baseline scenario, are shown in Table \ref{Cost-weight}.
\begin{table}[h]
	\caption{The results of the optimal human-like weight}
	\label{Cost-weight}
	\centering
	\begin{tabular}{c|cccc}
		\hline
		Scenarios      & $w_h$ & $w_r$ & $w_{\Gamma}$ & $w_s$ \\ \hline
		Base(Straight) & 0.184 & 0.377 & 0.167        & 0.272 \\
		Ramp           & 0.879 & 0.085 & 0.007        & 0.028 \\
		Cross          & 0.241 & 0.542 & 0.114        & 0.102 \\
		Long\_curve    & 0.010 & 0.019 & 0.875        & 0.097 \\
		Undulate       & 0.848 & 0.072 & 0.014        & 0.065 \\
		Rough          & 0.177 & 0.797 & 0.012        & 0.013 \\ \hline
	\end{tabular}
\end{table}

\subsubsection{Group 1: Straight and Ramp scenario}
\label{531}

\begin{figure}[h]
	\begin{minipage}{\textwidth}
		\begin{subfigure}{\textwidth}
			\centering
			\includegraphics[width=0.85\linewidth]{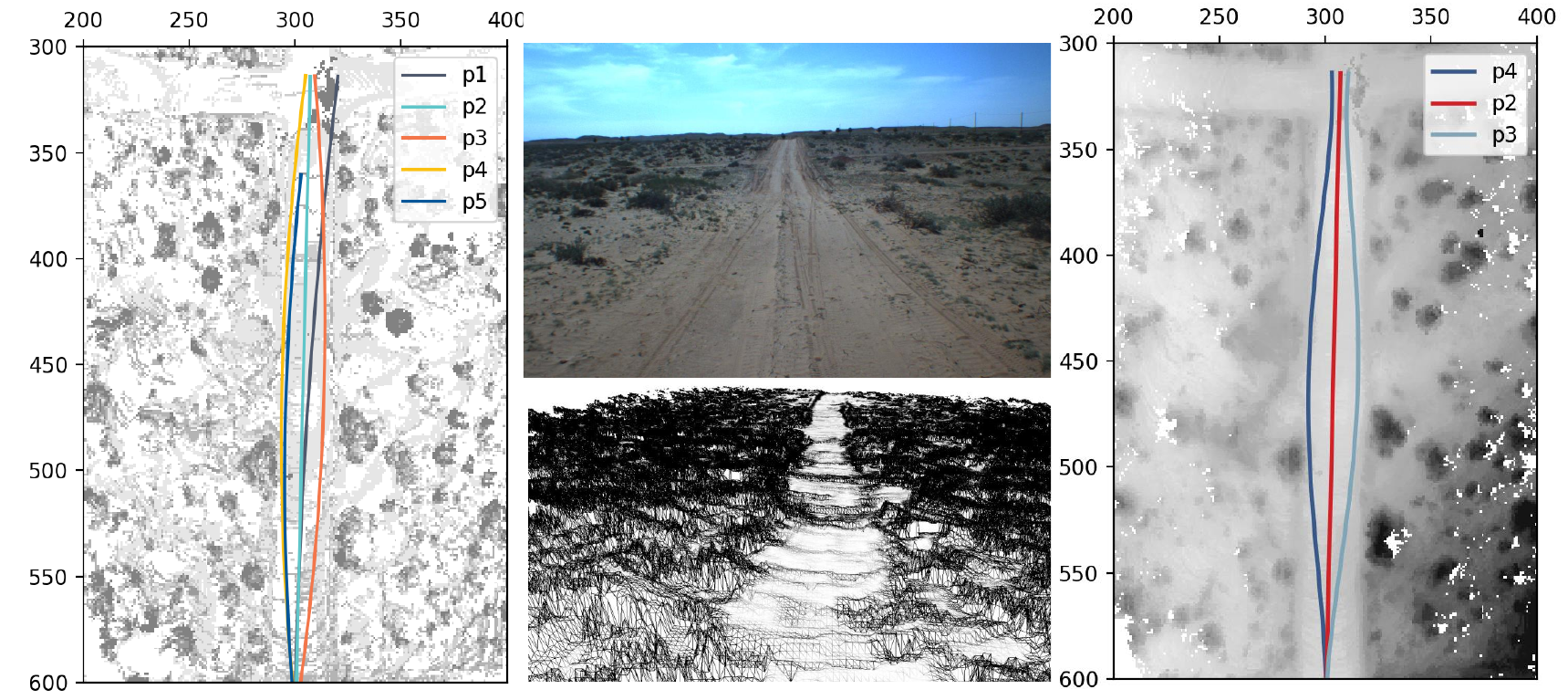}
			\captionsetup{font=footnotesize}
			\caption{The Straight (Base) Scenario. 'p1' - 'p5' are a subset of candidate trajectories generated by the motion planner. These trajectories in the left represent the optimal and part of the top choices according to the cost function formed by comparison weights in this group, and there are only optimal trajectories in the right. In the right image, the red line 'p2' is the optimal trajectory selected based on the optimal weight of the straight scenario, while 'p4' is the trajectory selected based on the optimal weight of the comparison group --- ramp scenario. It can be observed that the trajectory of 'p2' is more efficient than the others. }
			\label{Ra}
		\end{subfigure}
		\vfill % 增加子图之间的垂直间距
		\begin{subfigure}{\textwidth}
			\centering
			\includegraphics[width=0.85\linewidth]{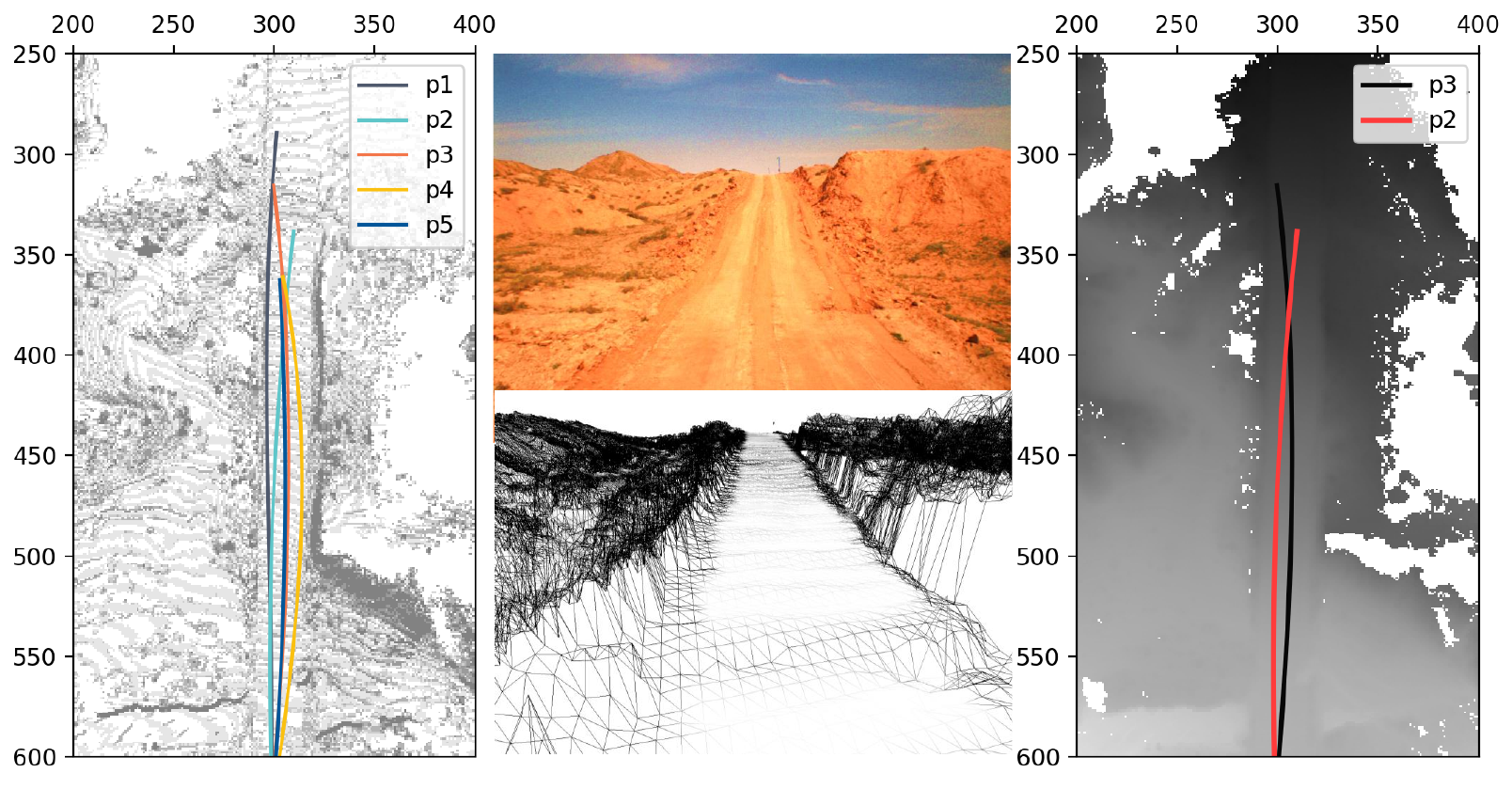}
			\captionsetup{font=footnotesize}
			\caption{The Ramp Scenario. Similar to  Fig.\ref{Ra}, in the right image, the red line 'p2' represents the optimal trajectory selected based on the ramp scenario, while 'p3' is selected based on the optimal weight of the straight scenario. It is noticed that 'p2' and 'p3' have quite different trajectory choices, with 'p2' choosing the left side of the ramp and having a slower speed compared to 'p3'.}
			\label{Rb}
		\end{subfigure}
		\caption{Group 1: The Straight(Base) Scenario and the Ramp Scenario. The layout of every whole image is: 1) The left image background is the roughness map, and lines are the candidate trajectories. 2) The top middle image is show the scenario, the bottom middle image is the 3D reconstruction by Lidar. 3) The right image background is the altitude map, and the red line representing the optimal trajectory.}
		\label{group1}	
	\end{minipage}
\end{figure}

The first group consisted of a straight road scenario (where the optimal cost weights in this scenario are chosen as the baseline) and a ramp scenario. The ramp scenario served as a comparison to validate that height is an important cost in some off-road scenarios even if the planning trajectories are straight from the top view. The planning results of both scenarios are shown in Fig.\ref{group1} . 

\begin{table}[h]
	\caption{Straight Scenario Candidate Trajectory Detailed Costs}
	\label{C_1}
	\centering
	\begin{tabular}{ccccc|cc}
		\hline
		&                           &                           &                           &                           & \multicolumn{2}{c}{Total Cost}                              \\
		\multirow{-2}{*}{plan} & \multirow{-2}{*}{cost\_H} & \multirow{-2}{*}{cost\_R} & \multirow{-2}{*}{cost\_T} & \multirow{-2}{*}{cost\_S} & Base                         & Ramp                         \\ \hline
		1                      & 0.546                     & 0.802                     & 0.980                     & 0.571                     & 0.704                        & 0.579                        \\
		2                      & 0.469                     & 0.510                     & 0.264                     & 0.411                     & {\color{blue} 0.358} & 0.461                        \\
		3                      & 0.485                     & 0.454                     & 1.481                     & 0.488                     & 0.673                        & 0.512                        \\
		4                      & 0.428                     & 0.448                     & 0.804                     & 0.627                     & 0.569                        & {\color{blue} 0.458} \\
		5                      & 0.474                     & 0.442                     & 0.962                     & 0.548                     & 0.585                        & 0.492                        \\ \hline
	\end{tabular}
\end{table}
\begin{table}[h]
	\caption{Ramp Scenario Candidate Trajectory Detailed Costs}
	\label{C_2}
	\centering
	\begin{tabular}{ccccc|cc}
		\hline
		&                                    &                                    &                                    &                                    & \multicolumn{2}{c}{{Total  Cost}}                    \\
		\multirow{-2}{*}{{traj\_i}} & \multirow{-2}{*}{{Cost\_H}} & \multirow{-2}{*}{{Cost\_R}} & \multirow{-2}{*}{{Cost\_T}} & \multirow{-2}{*}{{Cost\_S}} & line                         & ramp                         \\ \hline
		1                                  & 0.911                              & 0.519                              & 1.484                              & 0.253                              & 0.715                        & 0.842                        \\
		2                                  & 0.838                              & 0.444                              & 0.596                              & 0.535                              & 0.595                        & {\color{blue} 0.776} \\
		3                                  & 0.893                              & 0.474                              & 0.345                              & 0.298                              & {\color{blue} 0.491} & 0.795                        \\
		4                                  & 0.872                              & 0.546                              & 0.942                              & 0.755                              & 0.766                        & 0.841                        \\
		5                                  & 0.871                              & 0.434                              & 0.726                              & 0.305                              & 0.553                        & 0.786                        \\ \hline
	\end{tabular}
\end{table}
\begin{figure}[h]
	\centering
	\includegraphics[width=0.45\linewidth]{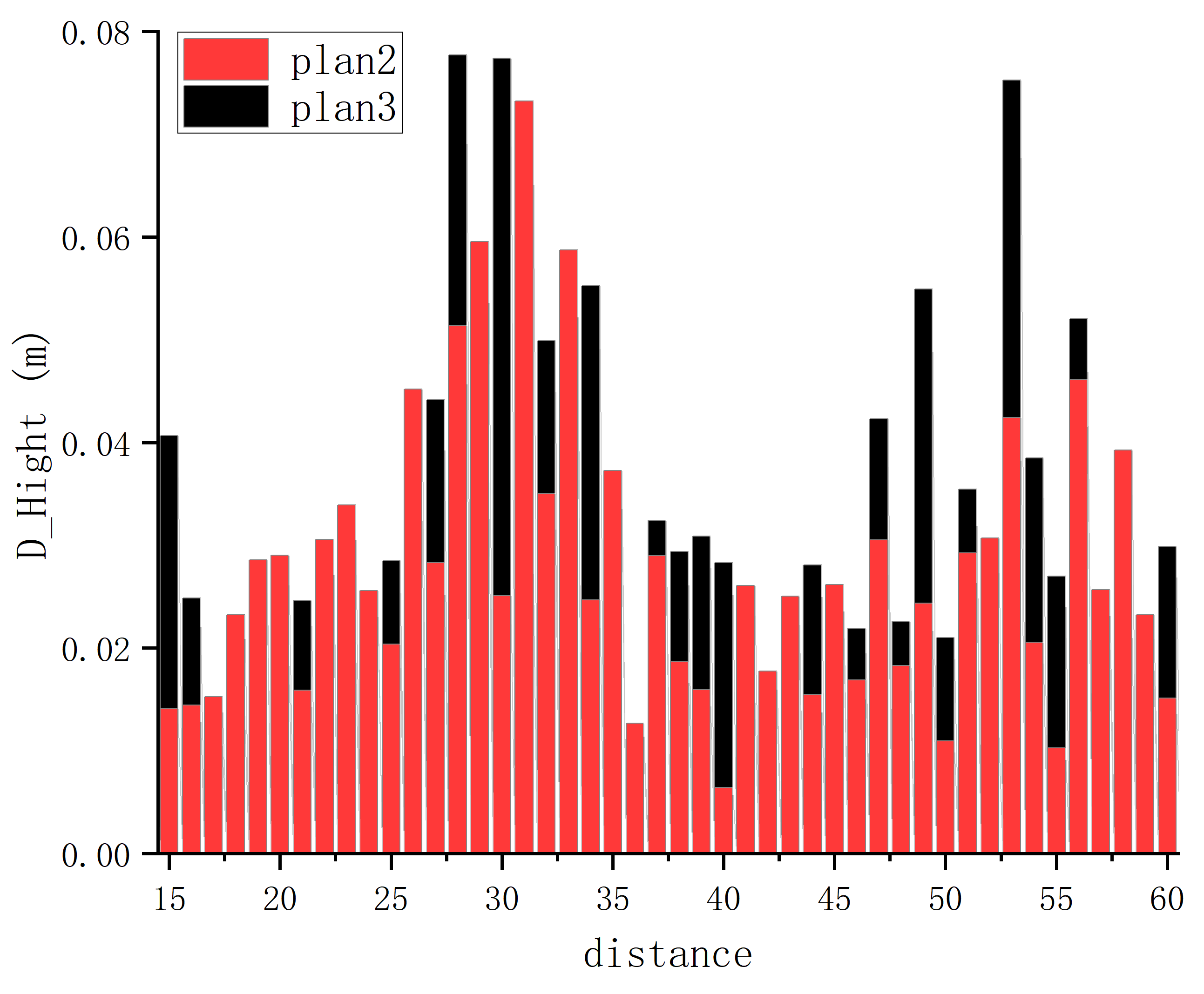}
	\caption{The height variation graph of the compared trajectories in Ramp scenario. The plot displays the rate of height change for each trajectory, with the black bars representing the height variation rate of baseline trajectory and the red bars representing the optimal trajectory.}
	\label{2_dh}
\end{figure}

The first and second rows of the Table.\ref{Cost-weight} show the optimal weights for these two scenarios based on human-like prediction results. It can be seen that the weights of each cost are basically consistent in the straight scenario. However, in the ramp scenario, the weight of the height cost is much higher than the other weights. Fig.\ref{Rb} shows that a naturally formed ramp in an off-road environment is relatively steep and uneven. Table.\ref{C_1} and \ref{C_2} show the detailed cost and total cost calculated by each optimal human-like weight. Fig.\ref{2_dh} shows the altitude and altitude variation of trajectories 'p2' and 'p3' in the ramp scenario. It shows that the altitude change in 'p2' is flatter than in 'p3'. 

The optimal weight assignments obtained from different scenarios learning from human cognition can lead to more appropriate planning results. The result of the experiment shows that human-like planning helps the motion planner to choose the trajectory that avoids steep slopes, which can significantly improve the mobility of the vehicle in the same ramp scenario. In addition, the slower speed in optimal 'p2' improves the vehicle's dynamics when climbing. Therefore, human-like planning and timely updating of planning strategies in response to environmental changes can effectively improve vehicle stability and dynamics.

\subsubsection{Group 2: Cross and Long-Curve Scenario}
\label{532}

\begin{figure}[h]
	\begin{minipage}{\textwidth}
		\begin{subfigure}{\textwidth}
			\centering
			\includegraphics[width=0.9\linewidth]{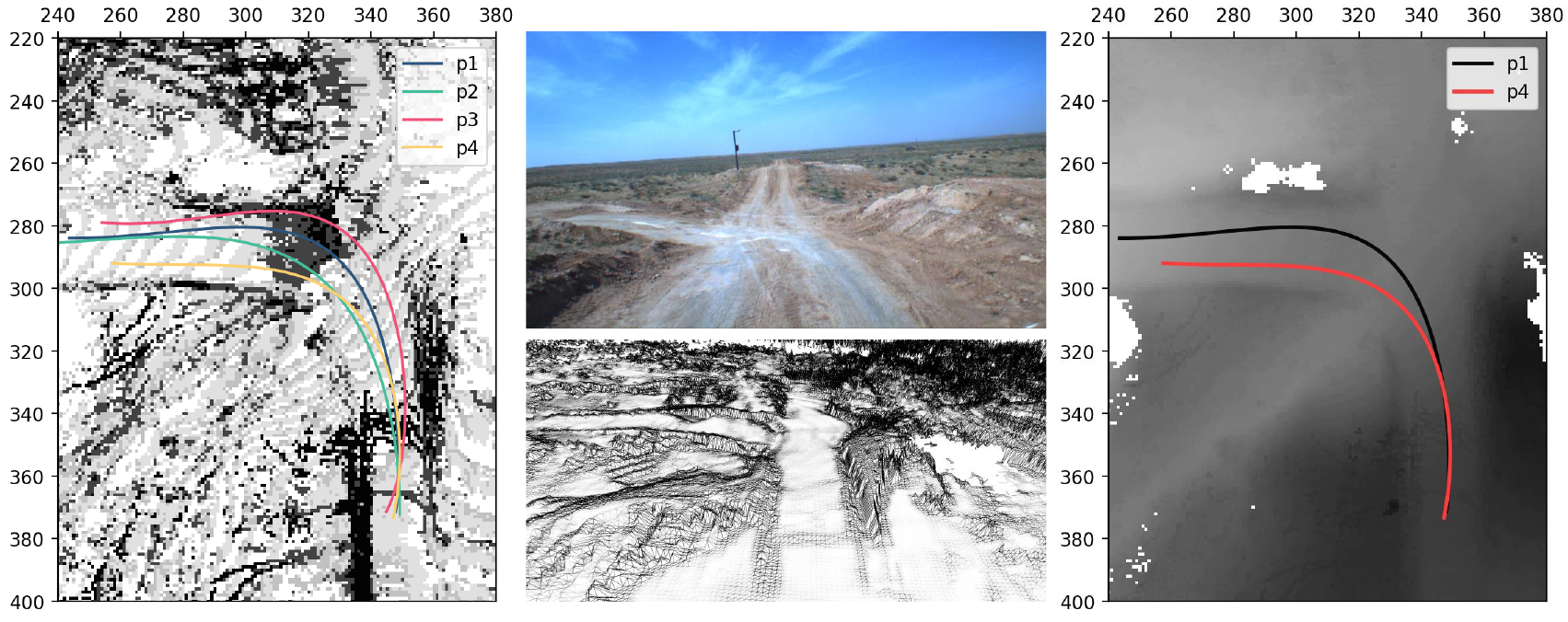}
			\captionsetup{font=footnotesize}
			\caption{The Cross Scenario. Similar to  Fig.\ref{Ra}, in the right image, the red line 'p4' is the optimal trajectory selected based on the cross scenario, while p1 is selected based on the optimal weight of the base scenario. The natural cross has complex terrain. 'p1' chooses a trajectory that leans towards the outer of the curve, while 'p4' leans towards the inner side of the curve and has a significantly slower speed than 'p1'.}
			\label{Rc}
		\end{subfigure}
		\vfill
		\begin{subfigure}{\textwidth}
			\centering
			\includegraphics[width=0.9\linewidth]{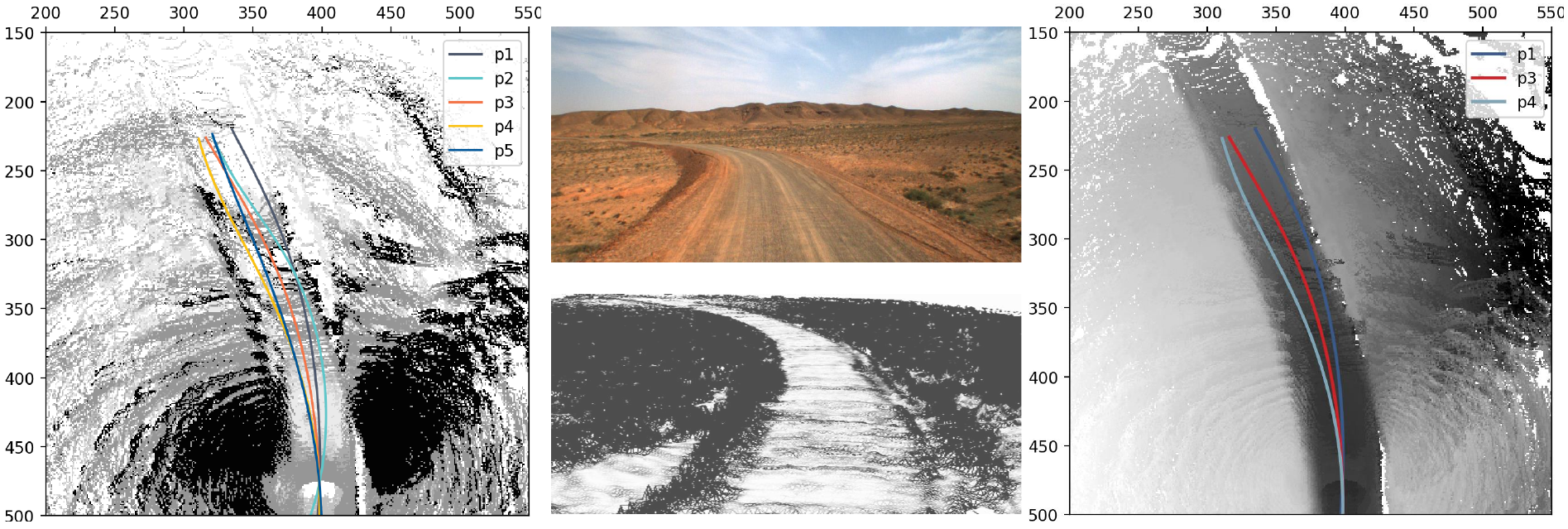}
			\captionsetup{font=footnotesize}
			\caption{The Long-curve Scenario. Similar to Fig.\ref{Ra}, in the right image, the red line 'p3' in the right image represents the optimal trajectory selected based on the long-curve scenario, while 'p1' is selected based on the optimal weight of the base scenario, 'p4' is selected based on the optimal weight of the comparison group --- cross scenario. The long-curved road with a smooth surface and is significantly higher than both sides. 'p3' stays away from the edge of the road, while both 'p1' and 'p4' have parts of their trajectories very close to the edge of the road, posing a risk of rollover.}
			\label{Rd}
		\end{subfigure}
		%\captionsetup{font=mysmall}
		\caption{Group 2: The Cross Scenario and the Long-curve Scenario. The layout of every whole image is similar to Figure \ref{group1}.}
		\label{group2}	
	\end{minipage}
\end{figure}

The second group consisted of a cross scenario with turning behavior and a long-curve scenario. The goal is to verify that the human-like weight emphasizes different factors in scenarios with similar driving behavior but different environmental factors. The planning results of both scenarios are shown in Fig.\ref{group2} . 

\begin{table}[h]
	\caption{Cross Scenario Candidate Trajectory Detailed Costs}
	\label{C_3}
	\centering
	\begin{tabular}{ccccc|cc}
		\hline
		&                           &                           &                           &                           & \multicolumn{2}{c}{Total Cost}                              \\
		\multirow{-2}{*}{plan} & \multirow{-2}{*}{cost\_H} & \multirow{-2}{*}{cost\_R} & \multirow{-2}{*}{cost\_T} & \multirow{-2}{*}{cost\_S} & Base                         & Cross                        \\ \hline
		1                      & 0.577                     & 0.813                     & 0.757                     & 1.488                     & {\color{blue} 0.960} & 0.782                        \\
		2                      & 0.548                     & 0.584                     & 0.606                     & 2.117                     & 1.064                        & 0.690                        \\
		3                      & 0.496                     & 0.931                     & 1.606                     & 2.213                     & 1.366                        & 0.883                        \\
		4                      & 0.566                     & 0.608                     & 0.785                     & 1.666                     & 0.967                        & {\color{blue} 0.676} \\ \hline
	\end{tabular}
\end{table}
\begin{table}[h]
	\caption{Long\_curve Scenario Candidate Trajectory Detailed Costs}
	\label{C_4}
	\centering
		\begin{tabular}{ccccc|ccc}
			\hline
			&                           &                           &                           &                           & \multicolumn{3}{c}{Total Cost}                                                                                       \\
			\multirow{-2}{*}{plan} & \multirow{-2}{*}{cost\_H} & \multirow{-2}{*}{cost\_R} & \multirow{-2}{*}{cost\_T} & \multirow{-2}{*}{cost\_S} & Base                         & Cross                        & \begin{tabular}[c]{@{}c@{}}Long\\ \_Curve\end{tabular} \\ \hline
			1                      & 0.708                     & 1.330                     & 1.262                     & 1.605                     & {\color{blue} 1.258} & 1.135                        & 1.369                                                  \\
			2                      & 0.893                     & 1.594                     & 1.425                     & 2.204                     & 1.589                        & 1.396                        & 1.704                                                  \\
			3                      & 0.754                     & 1.216                     & 0.690                     & 2.226                     & 1.324                        & 1.130                        & {\color{blue} 1.307}                           \\
			4                      & 0.748                     & 1.209                     & 1.316                     & 1.817                     & 1.313                        & {\color{blue} 1.096} & 1.470                                                  \\
			5                      & 0.826                     & 1.325                     & 2.112                     & 2.220                     & 1.644                        & 1.225                        & 2.040                                                  \\ \hline
		\end{tabular}
\end{table}

\begin{figure}[h]
	\begin{minipage}{\textwidth}
		\begin{minipage}{0.45\textwidth}
			\centering
			\includegraphics[width=\linewidth]{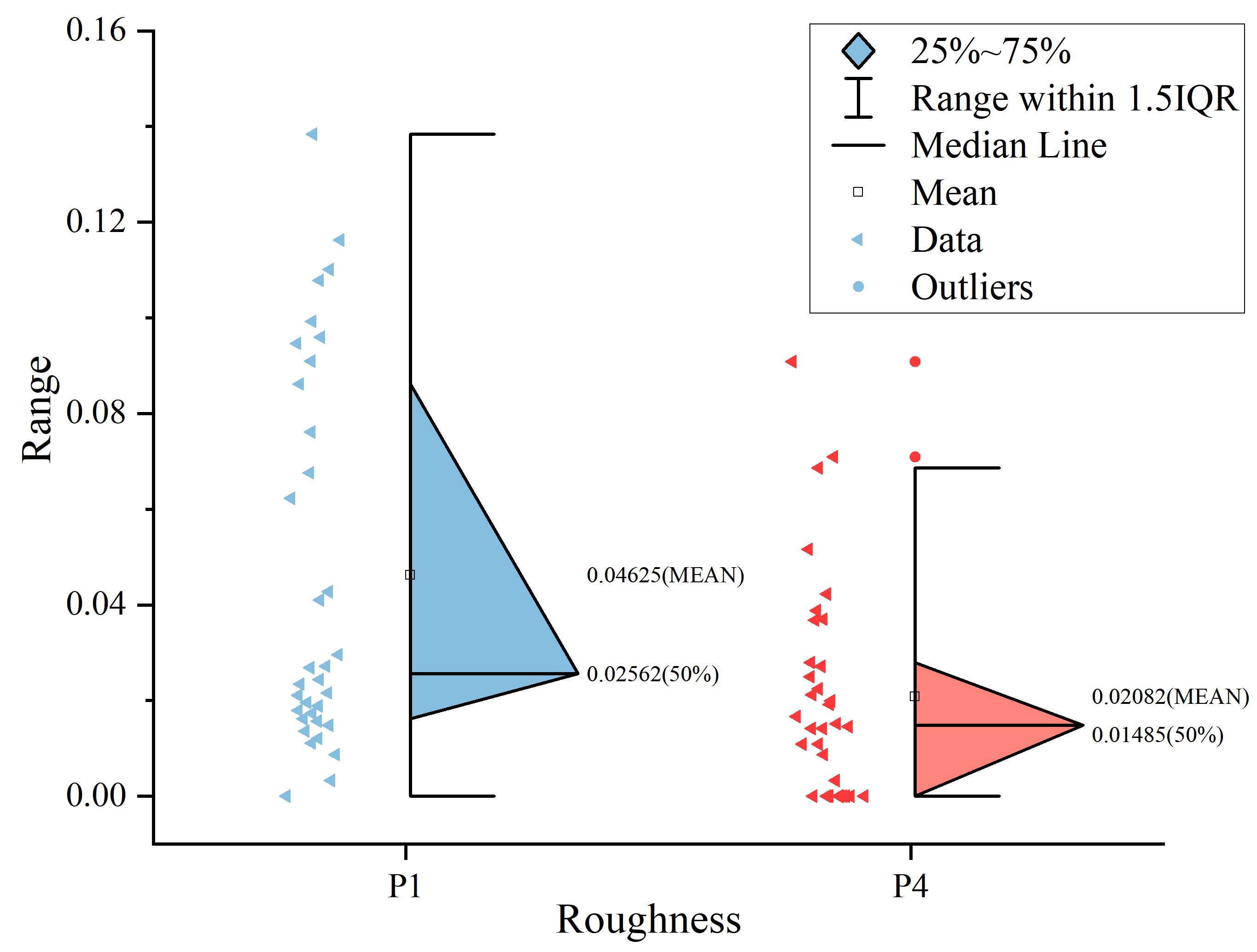}
		\end{minipage}
		\begin{minipage}{0.45\textwidth}
			\centering
			\includegraphics[width=\linewidth]{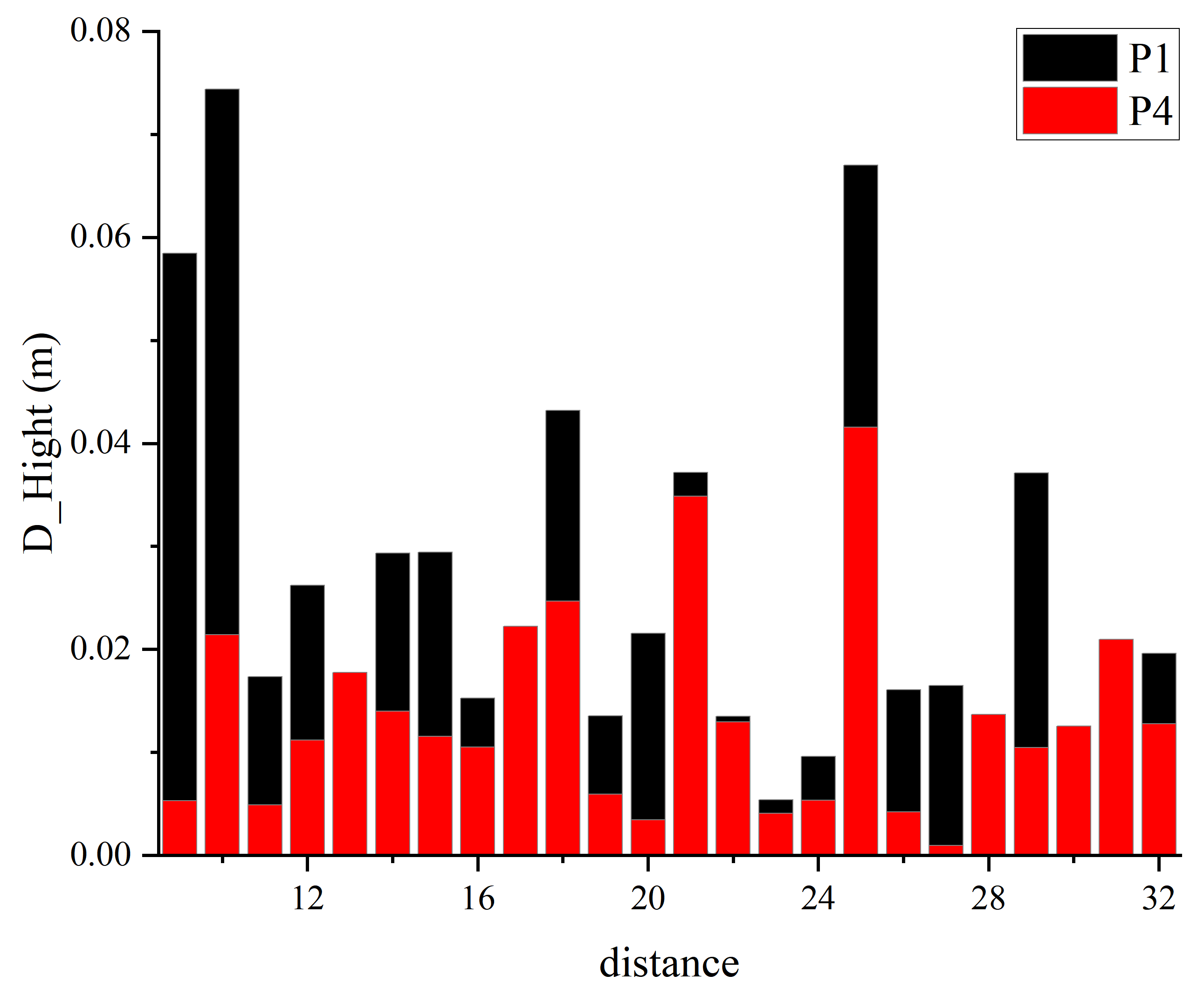}
		\end{minipage}
		\captionsetup{font=mysmall}
		\caption{The roughness and height variation graph of the compared trajectories in the cross scenario}
		\label{3_RH}
	\end{minipage}
\end{figure}

The third and fourth rows of Table.\ref{Cost-weight} show the optimal weights based on the human-like prediction results.  It can be seen that the weight in the cross scenario places more emphasis on the height and rough costs. Table. \ref{C_3} shows that the optimal planning trajectory is 'p1' if we still use the baseline weight. But if we use the human-like cost weight in the current, the actual optimal trajectory is 'p4'. Fig.\ref{3_RH} shows that 'p4' is better than 'p1', no matter the lower roughness or the more gradual height variation. In contrast, the long-curve scenario is less concerned with the cost of roughness and height, but rather emphasizes how to maintain proximity to the reference trajectory (i.e., the centerline of the lane).  Table. \ref{C_4} shows that the 'p3' is the optimal trajectory if we use the calculated weights in this scenario, and it is obviously better than others.  'p4', which we actually choose, is to always stay away from the edges of the road where it is likely to roll over.

Even for similar driving behaviors, drivers may pay attention to different environmental factors.  The result in this cross scenario indicates that human drivers focus on the terrain if the cross terrain is complex enough, which can improve stability during turning behavior. But in this long curve scenario if driving too close to the side of the road is likely to result in a rollover. Thus, the human-like planner places more emphasis on trajectory deviation cost(staying closer to the road centerline) than others. This comparison illustrates that even for scenarios with similar driving behaviors, the weighting of each cost can be significantly different. Therefore, finding the optimal weights specific to the current context is more favorable for vehicles in different off-road scenarios.

\subsubsection{Group 3: Undulate and Rough Scenario}
\label{533}

\begin{figure}[h]
	\begin{minipage}{\textwidth}
		\vfill
		\begin{subfigure}{\textwidth}
			\centering
			\includegraphics[width=0.85\linewidth]{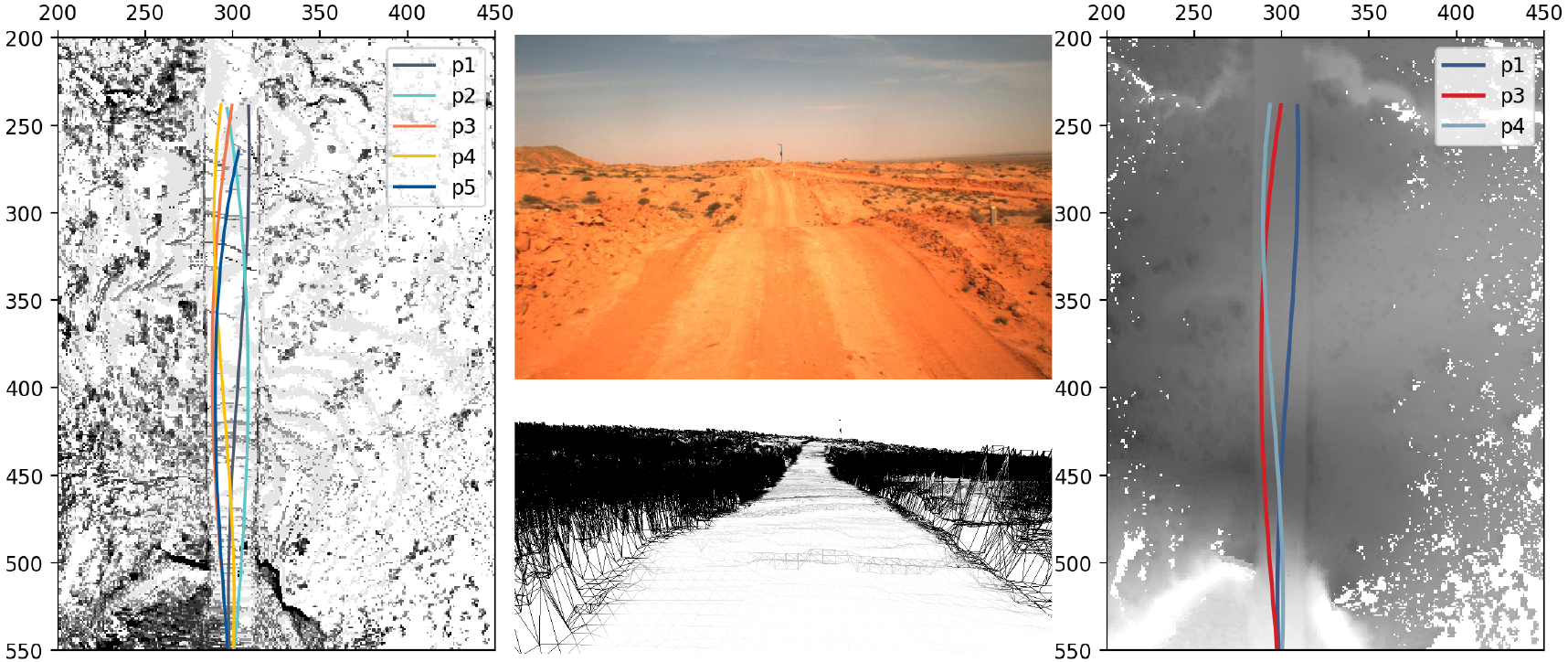}
			\captionsetup{font=footnotesize}
			\caption{The Undulate Scenario. Similar to Fig.\ref{Ra}, in the right image, the red line 'p3' represents the optimal trajectory selected based on the undulate scenario, while 'p1' is selected based on the optimal weight of the base scenario. 'p4' is selected based on the optimal weight of the comparison group --- rough scenario. The image shows a road with complex undulations. The optimal trajectories selected based on different scenarios show significant variations, with 'p3' favoring the left side of the road.}
			\label{Re}
		\end{subfigure}
		\vfill
		\begin{subfigure}{\textwidth}
			\centering
			\includegraphics[width=0.85\linewidth]{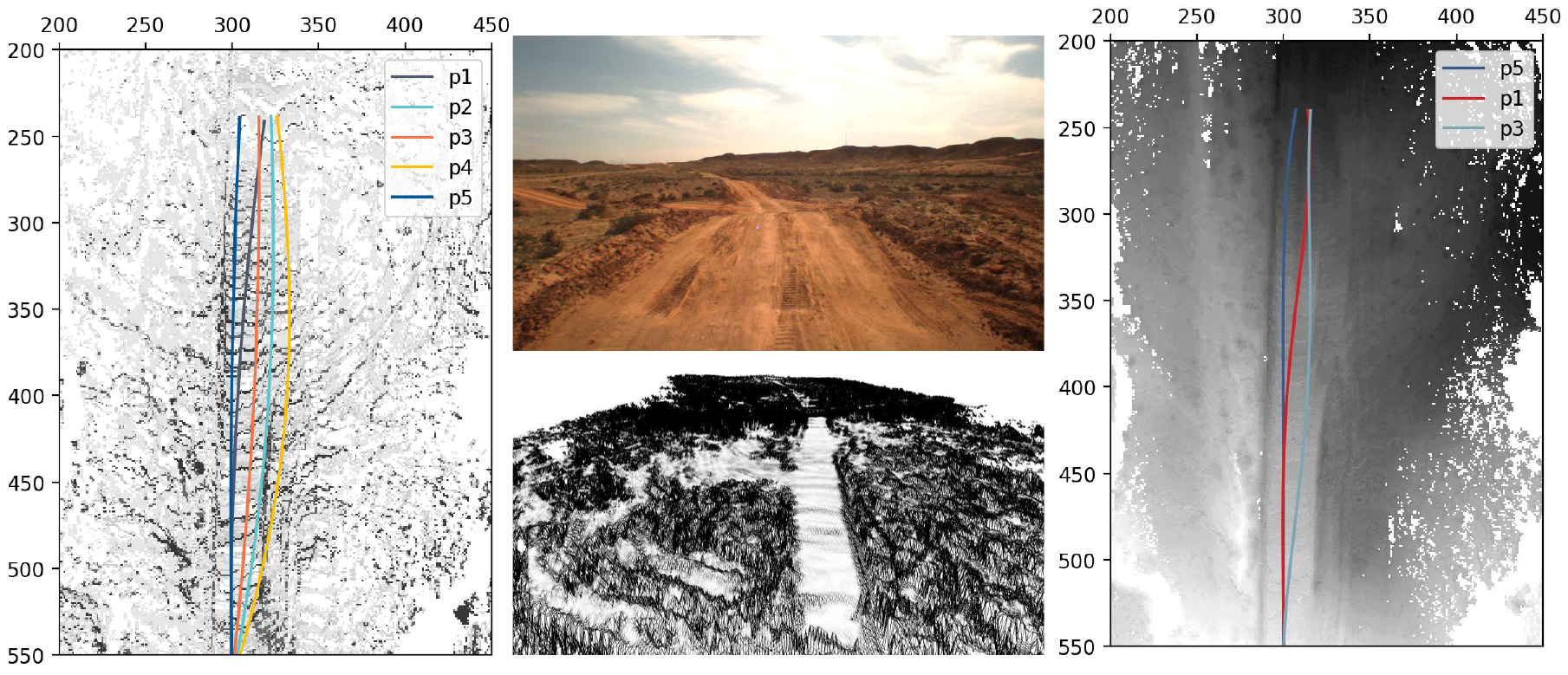}
			\captionsetup{font=footnotesize}
			\caption{The Rough Scenario. Similar to Fig.\ref{Ra}, in the right image, the red line 'p1' represents the optimal trajectory selected based on the rough scenario, while 'p5' is selected based on the optimal weight of the base scenario. 'p3' is based on the optimal weight of the comparison group --- undulate scenario. The image shows a road with significant roughness. The optimal trajectory selected based on the different scenarios shows significant differences, and the optimal 'p1' appears to be more tortuous from the top view only.}
			\label{Rf}
		\end{subfigure}	
		\captionsetup{font=mysmall}
		\caption{Group 3: The Undulate Scenario and the Rough Scenario. The layout of every whole image is similar to Figure \ref{group1}.}
		\label{group3}	
	\end{minipage}
\end{figure}

The third group consisted of an undulated scenario and a rough scenario. The goal is to verify that the human-like weight focuses more specifically on the most significant environmental costs in the current scenarios with different primary terrain features. The planning results of both scenarios are shown in Fig.\ref{group3} .

\begin{figure}[h]
	\begin{minipage}{\textwidth}
		\begin{subfigure}{0.45\textwidth}
			\centering
			\includegraphics[width=\linewidth]{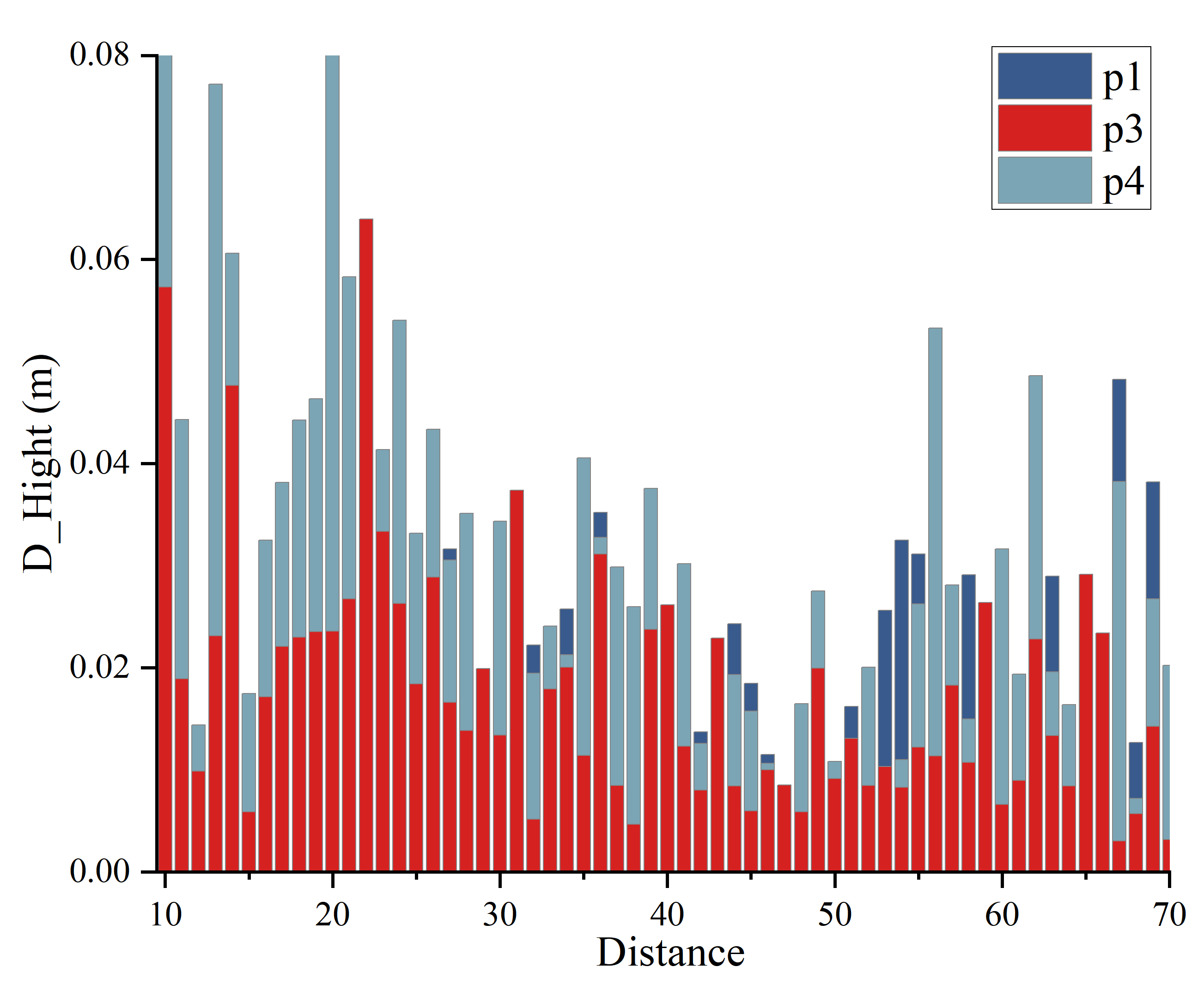}
			\captionsetup{font=footnotesize}
			\caption{The height variation graph of the compared trajectories in undulate scenario}
			\label{5_H}
		\end{subfigure}
		\begin{subfigure}{0.45\textwidth}
			\centering
			\includegraphics[width=\linewidth]{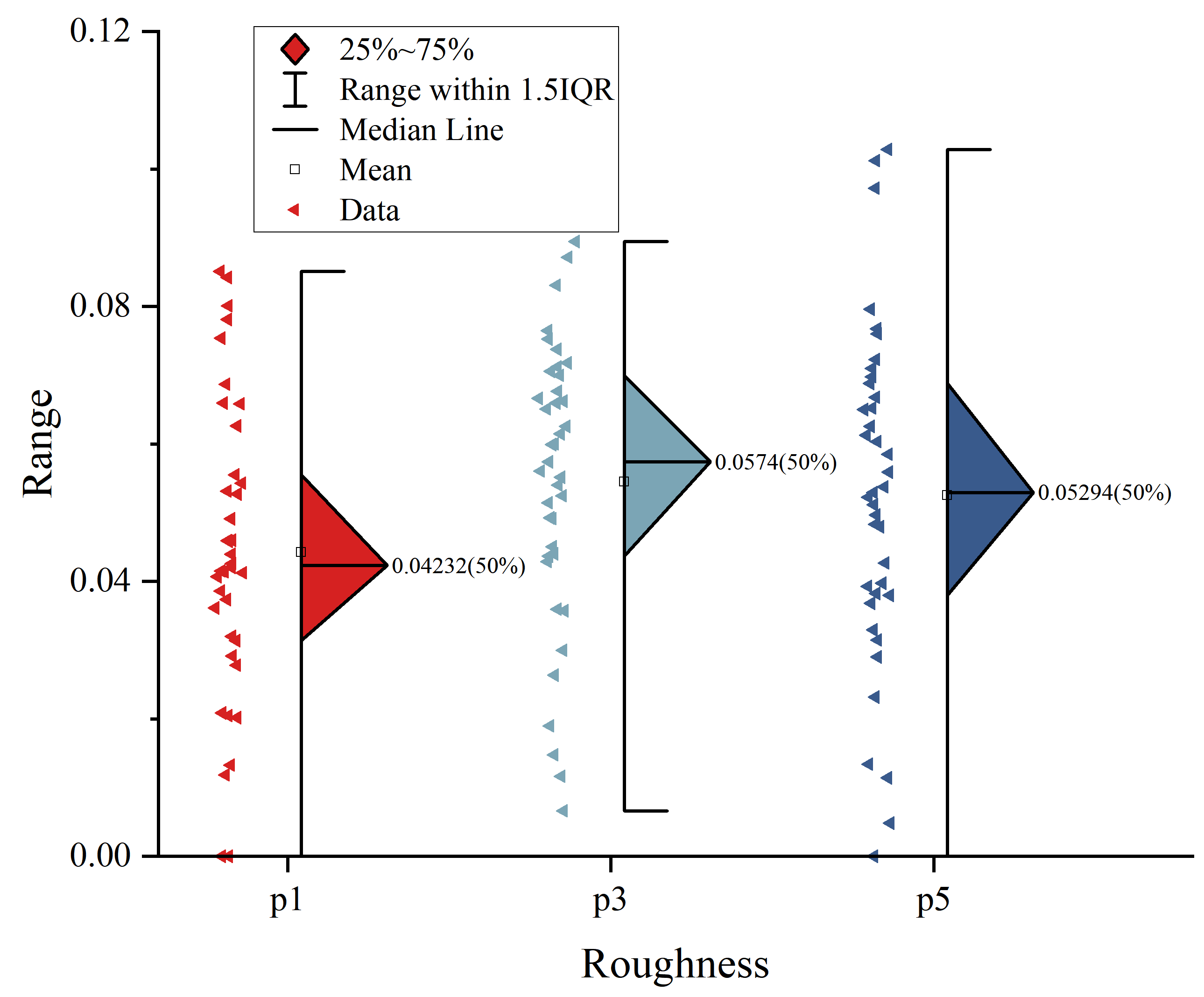}
			\captionsetup{font=footnotesize}
			\caption{The roughness graph of the compared trajectories in rough scenario}
			\label{6_R}
		\end{subfigure}
		\captionsetup{font=mysmall}
		\caption{The performance in undulate and rough scenario }
	\end{minipage}
\end{figure}
\begin{table}[h]
	\caption{Undulate Scenario Candidate Trajectory Detailed Costs}
	\label{C_5}
	\centering
		\begin{tabular}{ccccc|ccc}
			\hline
			&                           &                           &                           &                           & \multicolumn{3}{c}{Total Cost}                                                             \\
			\multirow{-2}{*}{plan} & \multirow{-2}{*}{cost\_H} & \multirow{-2}{*}{cost\_R} & \multirow{-2}{*}{cost\_T} & \multirow{-2}{*}{cost\_S} & Base                         & Undulate                     & Rough                        \\ \hline
			1                      & 0.638                     & 0.510                     & 0.786                     & 0.431                     & {\color{blue} 0.569} & 0.604                        & 0.535                        \\
			2                      & 0.647                     & 0.544                     & 0.962                     & 0.594                     & 0.666                        & 0.630                        & 0.568                        \\
			3                      & 0.621                     & 0.498                     & 1.426                     & 0.488                     & 0.705                        & {\color{blue} 0.600} & 0.531                        \\
			4                      & 0.655                     & 0.459                     & 0.981                     & 0.492                     & 0.617                        & 0.614                        & {\color{blue} 0.500} \\
			5                      & 0.673                     & 0.531                     & 1.160                     & 0.441                     & 0.659                        & 0.638                        & 0.563         \\ \hline
		\end{tabular}
\end{table}
\begin{table}[h]
	\caption{Rough Scenario Candidate Trajectory Detailed Costs}
	\label{C_6}
	\centering
		\begin{tabular}{ccccc|ccc}
			\hline
			&                           &                           &                           &                           & \multicolumn{3}{c}{Total Cost}                                                             \\
			\multirow{-2}{*}{plan} & \multirow{-2}{*}{cost\_H} & \multirow{-2}{*}{cost\_R} & \multirow{-2}{*}{cost\_T} & \multirow{-2}{*}{cost\_S} & Base                         & Undulate                     & Rough                        \\ \hline
			1                      & 0.913                     & 0.985                     & 0.616                     & 0.342                     & 0.693                        & 0.882                        & {\color{blue} 0.959} \\
			2                      & 0.908                     & 1.243                     & 2.086                     & 0.519                     & 1.101                        & 0.950                        & 1.184                        \\
			3                      & 0.822                     & 1.039                     & 0.910                     & 0.297                     & 0.729                        & {\color{blue} 0.823} & 0.989                        \\
			4                      & 0.840                     & 1.030                     & 3.406                     & 1.027                     & 1.449                        & 0.913                        & 1.026                        \\
			5                      & 0.858                     & 1.019                     & 0.920                     & 0.162                     & {\color{blue} 0.692} & 0.837                        & 0.978                        \\ \hline
		\end{tabular}
\end{table}

The fifth and sixth rows of Table.\ref{Cost-weight} show the optimal weights based on the human-like prediction results. In the undulate scenario, drivers tend to focus more on changes in road elevation, and the weight given to the cost of height is significantly higher than other cost factors.  Conversely, in the rough scenario, drivers tend to prioritize the roughness of the road surface, with the weight given to the cost of roughness being significantly higher than other cost factors.  Table. \ref{C_5} and \ref{C_6} show that the optimal planning trajectory is 'p3' and 'p1' respectively with the human-like cost weight in the current. Fig.\ref{5_H} shows that the height variation of 'p3' is more gradual than other trajectories using other scenario weights.  Similarly in Fig.\ref{6_R}, the roughness of 'p1' is the lowest.

Both undulate and rough scenarios have factors that significantly affect vehicle motion. It is more stable for moving vehicles to choose flatter trajectories on undulating roads and less bumpy trajectories on rough roads. This has a significant impact on the adaptability to the environment and the stability of vehicle driving in autonomous off-road scenarios.

\subsubsection{The Statistical Results at Various Locations within Each Scenario }\label{534} 

\begin{figure}[h]
	\begin{minipage}{\textwidth}
		\begin{minipage}{0.5\textwidth}
			\includegraphics[width=\linewidth]{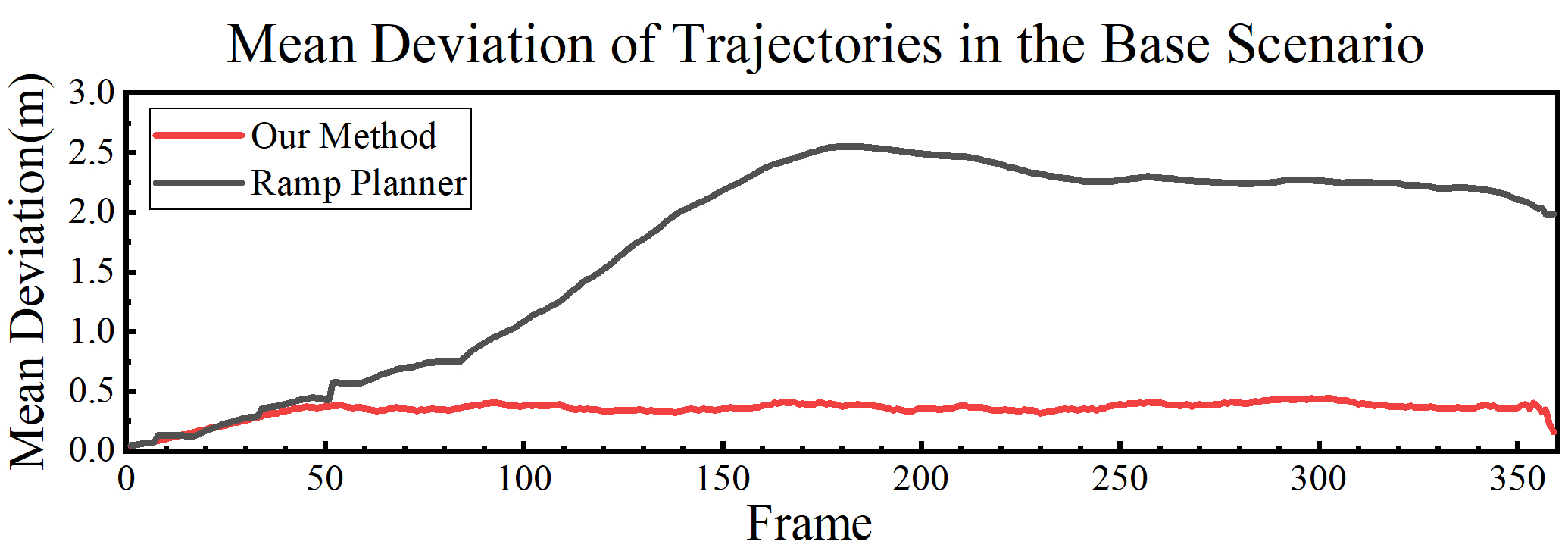}
		\end{minipage}
		\begin{minipage}{0.5\textwidth}
			\includegraphics[width=\linewidth]{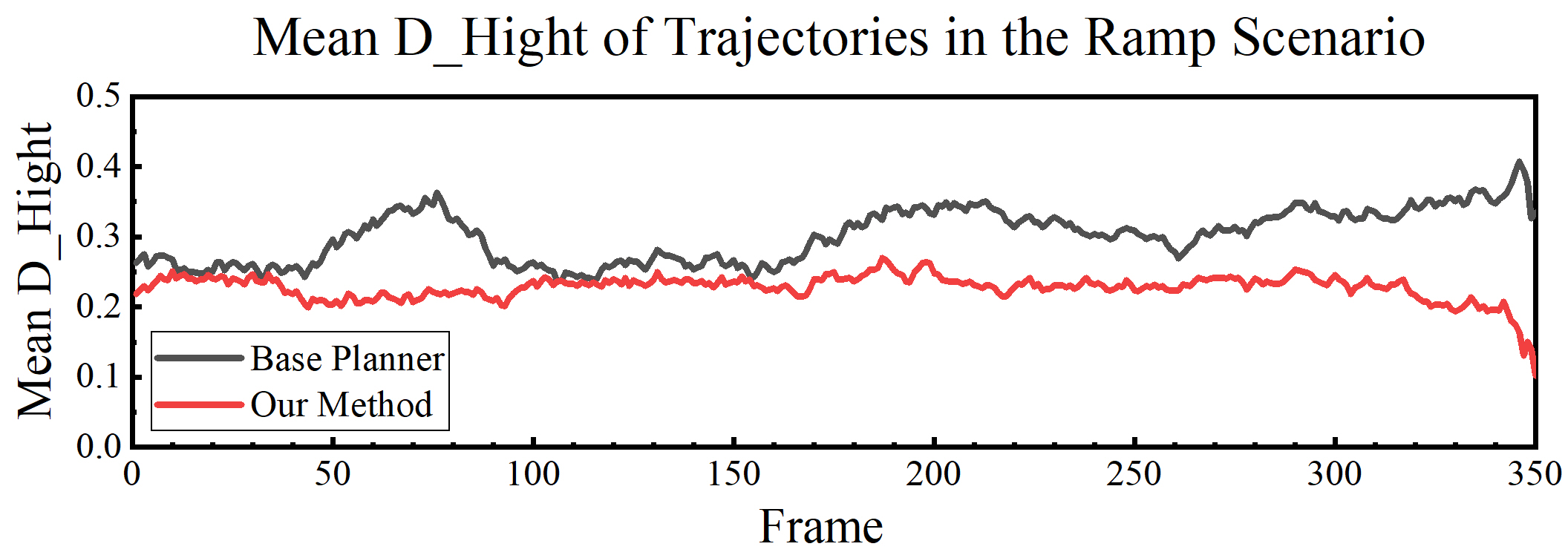}
		\end{minipage}
		\begin{minipage}{0.5\textwidth}
			\includegraphics[width=\linewidth]{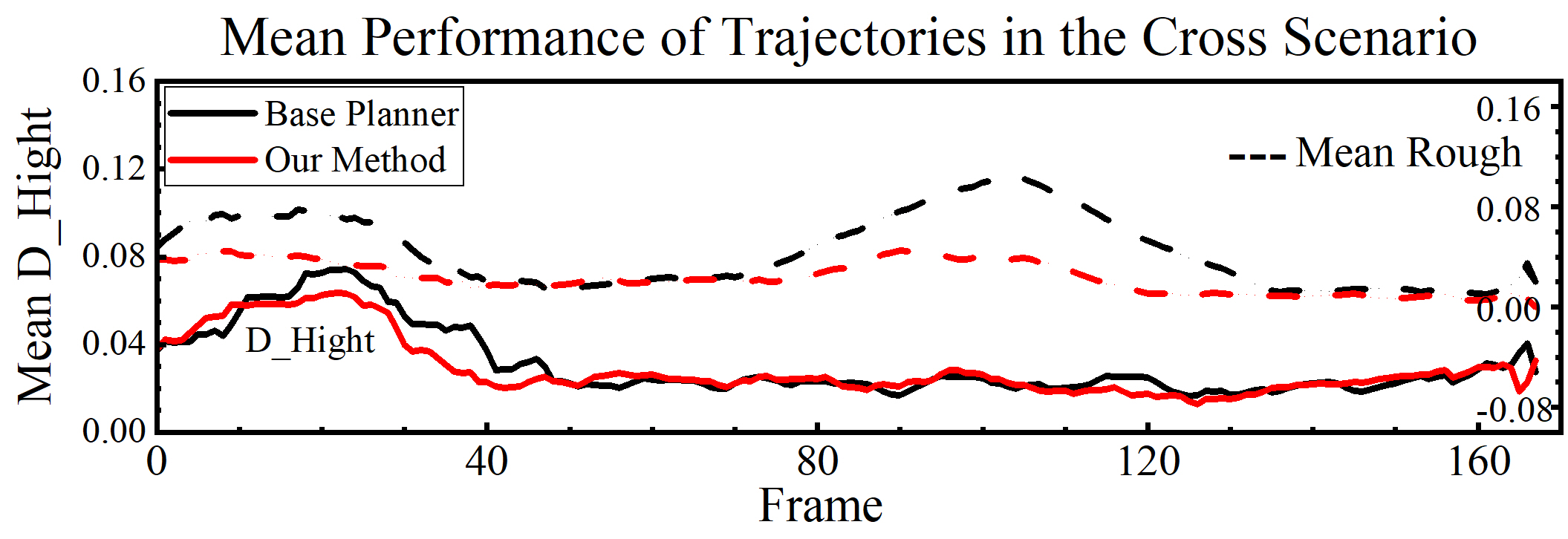}
		\end{minipage}
		\hfill
		\begin{minipage}{0.5\textwidth}
			\includegraphics[width=\linewidth]{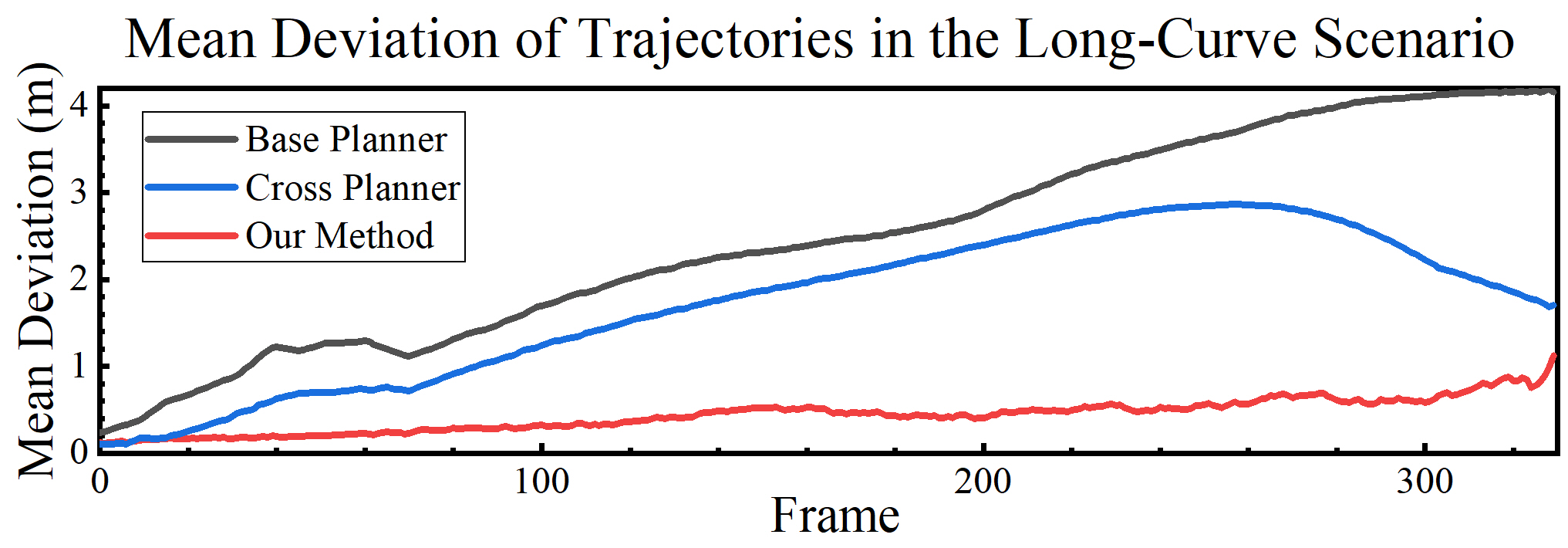}
		\end{minipage}
		\vfill
		\begin{minipage}{0.5\textwidth}
			\includegraphics[width=\linewidth]{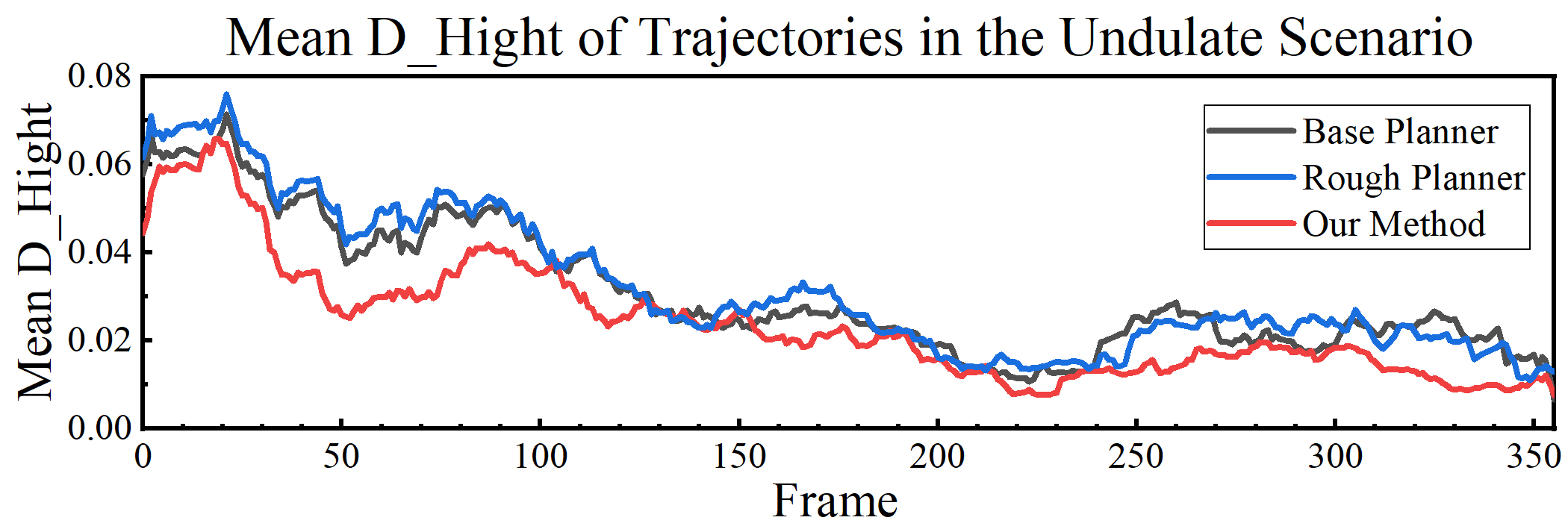}
		\end{minipage}
		\hfill
		\begin{minipage}{0.5\textwidth}
			\includegraphics[width=\linewidth]{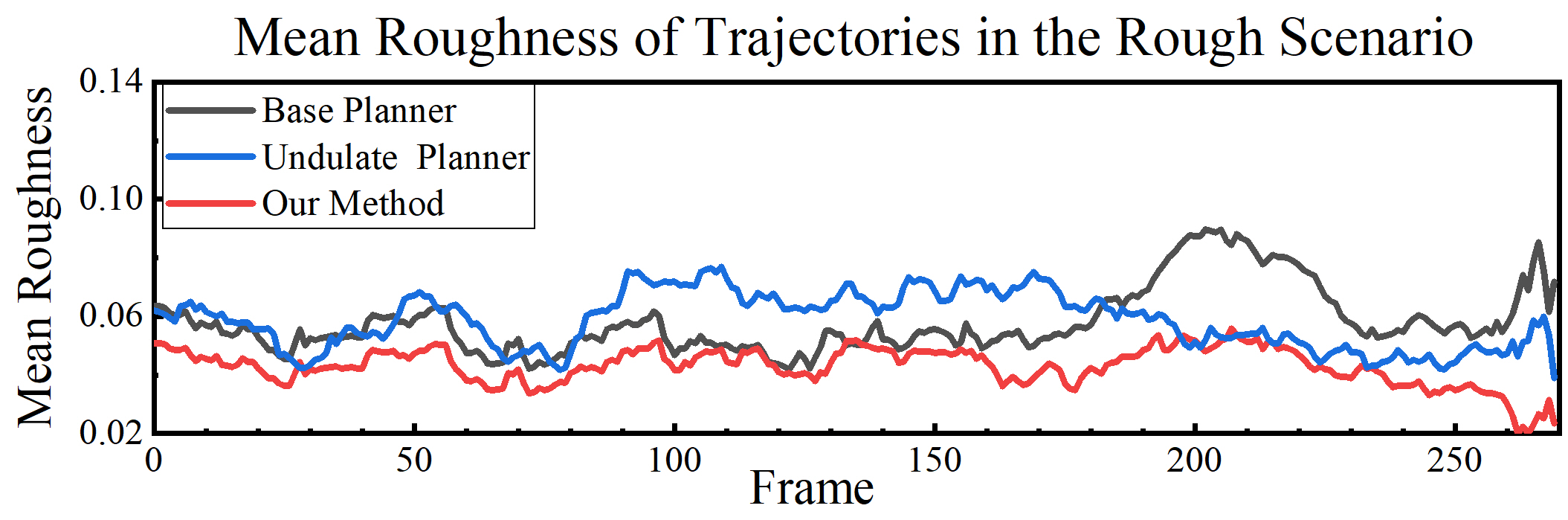}
		\end{minipage}
		\captionsetup{font=mysmall}
		\caption{Performance Evaluation of Trajectory Generation by Proposed and Compared planners in Aforementioned Scenarios. The red lines represent the trajectories generated by our method, while the blue and black lines represent the trajectories generated by compared fixed weights method. The values on the vertical axis (y-axis) in the figures represent the average performance of all the points on each planning trajectory generated at every frame (x-axis) during on-road vehicle testing. In the y-axis, 'Mean Deviation' represents the mean offset of the trajectories relative to the reference trajectory;'Mean D\_Hight' is the mean value of the road surface height variation that the trajectories pass through; 'Mean Rough' indicates the mean roughness of the road surface that the trajectories traverse.}
		\label{all_pic}	
	\end{minipage}
\end{figure}

The proposed method has evaluated every frame location within each scenario mentioned above. The corresponding statistical results are presented in Table.\ref{all_result} and Fig.\ref{all_pic}, demonstrating the most significant performance values within the current scenario. These values are the evaluations of the trajectories generated by our method and other compared fixed-weight planners that are being compared.

\begin{table}[h]
	\caption{The Statistical Results at Various Locations within Each Scenario."Optimal Selection" indicates that our method is the only one that generated the optimal trajectory, "Joint Optimal Selection" indicates that the comparative methods also generated, "Our Method is the optimal" is the sum of these two.}
	\centering
	\label{all_result}
	\begin{tabular}{ccccc}
		\hline
		Scenarios                                         & Our Method              & Test Count & \multicolumn{2}{c}{Percentage}                                \\ \hline
		\multirow{4}{*}{Base}         & Optimal Selection       & 347        & \multicolumn{2}{c}{96.657\%}                                  \\
		& Joint Optimal Selection & 5          & \multicolumn{2}{c}{1.393\%}                                   \\
		& Non-optimal Selection   & 7          & \multicolumn{2}{c}{1.950\%}                                   \\
		& total                   & 359        & Our Method is the optimal: & 98.050\%					   \\ \hline
		\multirow{4}{*}{Ramp}         & Optimal Selection       & 332        & \multicolumn{2}{c}{94.857\%}                                  \\
		& Joint Optimal Selection & 17         & \multicolumn{2}{c}{4.857\%}                                   \\
		& Non-optimal Selection   & 1          & \multicolumn{2}{c}{0.286\%}                                   \\
		& total                   & 350        & Our Method is the optimal: & 99.714\%                      \\ \hline
		\multirow{4}{*}{Cross}        & Optimal Selection       & 117        & \multicolumn{2}{c}{69.643\%}                                  \\
		& Joint Optimal Selection & 46         & \multicolumn{2}{c}{27.381\%}                                  \\
		& Non-optimal Selection   & 5          & \multicolumn{2}{c}{2.976\%}                                   \\
		& total                   & 168        & Our Method is the optimal: & 97.023\%                      \\ \hline
		\multirow{4}{*}{Long-Curve} & Optimal Selection       & 324        & \multicolumn{2}{c}{97.297\%}                                  \\
		& Joint Optimal Selection & 4          & \multicolumn{2}{c}{1.201\%}                                   \\
		& Non-optimal Selection   & 5          & \multicolumn{2}{c}{1.502\%}                                   \\
		& total                   & 333        & Our Method is the optimal: & 98.498\%                      \\ \hline
		\multirow{4}{*}{Undulate}         & Optimal Selection       & 322        & \multicolumn{2}{c}{89.694\%}                                  \\
		& Joint Optimal Selection & 22         & \multicolumn{2}{c}{6.128\%}                                   \\
		& Non-optimal Selection   & 15         & \multicolumn{2}{c}{4.178\%}                                   \\
		& total                   & 359        & Our Method is the optimal: & 95.822\%                      \\ \hline
		\multirow{4}{*}{Rough}        & Optimal Selection       & 260        & \multicolumn{2}{c}{95.588\%}                                  \\
		& Joint Optimal Selection & 7          & \multicolumn{2}{c}{2.574\%}                                   \\
		& Non-optimal Selection   & 5          & \multicolumn{2}{c}{1.838\%}                                   \\
		& total                   & 272        & Our Method is the optimal: & 98.162\%                      \\ \hline
	\end{tabular}
\end{table}

In Table.\ref{all_result}, we provide five descriptive statistics for the testing results of each scenario. Taking the Ramp Scenario as an example, the key metric for evaluating trajectory quality in the Ramp Scenario is the gradient, which refers to the minimum height variation of the generated trajectories, indicating the optimal trajectory. In the table, "Optimal Selection" indicates that our method is the only one that generated the optimal trajectory for this scenario, with a statistical value of 332 times, accounting for 94.857\% of the total number of tests. "Joint Optimal Selection" indicates that in addition to our method, the comparative methods also generated the optimal trajectory, with a statistical value of 17 times, accounting for 4.857\%. "Non-optimal Selection" indicates the statistical value of the times our method did not generate the optimal trajectory, which is 7 times, accounting for 0.286\%. "Total" represents the total number of 350 planning test results generated for this scenario. The probability of the trajectory generated by our method being the optimal choice is 99.714\%.

The table.\ref{all_result} demonstrates that our method consistently achieves optimal solutions of 95\% or higher. Furthermore, the figure depicts the sustained superiority of the trajectories produced by our method throughout the entirety of the testing process. These findings highlight the practical applicability of our approach.

The performance of trajectory generation by our proposed method and the compared planners was evaluated in the aforementioned six scenarios in Fig.\ref{all_pic}. We compared the significant performance of our proposed method's planned trajectories with those generated by other planners in the current scenario. The red color represents the trajectories generated by our method. Throughout all scenarios, our proposed method consistently exhibited superior performance in terms of trajectory generation throughout the testing process.

The aforementioned experiments provide compelling evidence that learning the human driver's perception of the environment in different driving scenarios and integrating it into the planning process can significantly improve the adaptability of planned trajectories in different scenarios. Our motion planner enables the generation of more stable and efficient trajectories. In addition, the adoption of the planner based on behavioral primitives facilitates a more comprehensive consideration of vehicle dynamics.

\section{Conclusion}
In this study, we proposed an adaptive motion planner for off-road driving, integrating human-like cognition and cost evaluation to generate optimal trajectories in complex and varied terrains. The experiments were conducted to validate the effectiveness of the proposed method and investigate its performance in diverse scenarios. The experiments were grouped into different scenarios to evaluate the adaptability and stability of the human-like motion planner. The results demonstrated the following key findings:

\begin{enumerate}
	\item{The optimal weight assignments obtained from different scenarios, learning from human cognition, led to more appropriate planning results. Human-like planning effectively avoided steep slopes in ramp scenarios, improving vehicle mobility and dynamics.}
	\item{Even for scenarios with similar driving behaviors, the weighting of each cost varied significantly. The human-like planner emphasized different factors based on the environmental characteristics, such as focusing on terrain in complex cross scenarios and prioritizing trajectory deviation in long-curve scenarios.}
	\item{The experiments in undulated and rough scenarios showed that the human-like weight prioritized significant environmental costs, resulting in more stable paths on undulating roads and less bumpy paths on rough terrains. This enhanced the adaptability and stability of the planned trajectories.}
\end{enumerate}

The empirical results provide convincing evidence that incorporating human driver's perception of the environment and integrating it into the planning process significantly improves the adaptability of planned trajectories in different off-road scenarios. The proposed motion planner generates more stable and efficient trajectories while considering vehicle dynamics.
However, this work has some limitations, both in terms of theoretical constraints and experimental observations. The theoretical limitation lies in the reliance on human-like cognition, which may not capture all possible driving strategies or account for unforeseen scenarios. Additionally, the experiments were conducted in specific off-road environments, which may not fully represent the complexity and variability of all off-road scenarios. To address these limitations, future work should focus on the following aspects:

\begin{enumerate}
	\item{Further explore and refine the theoretical framework by incorporating more diverse human driving strategies and considering a broader range of off-road scenarios.}
	\item{Extend the experiments to a wider variety of off-road environments with different terrains and road conditions to evaluate the robustness and generalizability of the proposed method.}
	\item{Investigate the potential of leveraging real-time sensory data and environmental feedback to enhance the adaptability and responsiveness of the motion planner.}
	\item{Further refinement of existing planning costs will be undertaken, along with the inclusion of additional costs relevant to planning, aiming to enhance planning quality.}
\end{enumerate}

By addressing these limitations and advancing the research, we can enhance the capabilities of autonomous vehicles in off-road environments, making them more capable of navigating through complex and varied terrains while ensuring stable and efficient travel.

\bibliography{ref}

\end{document}